\documentclass{article}

\usepackage{arxiv}

\usepackage[utf8]{inputenc} 
\usepackage[T1]{fontenc}    
\usepackage{hyperref}       
\usepackage{url}            
\usepackage{booktabs}       
\usepackage{amsfonts}       
\usepackage{nicefrac}       
\usepackage{microtype}      
\usepackage{lipsum}		
\usepackage{graphicx}
\usepackage{natbib}
\usepackage{doi}

\usepackage{array}
\usepackage{pdflscape}

\usepackage{caption}
\usepackage{subcaption}
\usepackage{pdflscape}
\usepackage{multirow}
\graphicspath{ {./images/} }
\usepackage{amsmath}
\usepackage{amssymb}
\usepackage{algorithm}
\usepackage{mathtools}
\usepackage{soul}
\usepackage[dvipsnames]{xcolor}
\usepackage{hyperref}
\usepackage{lineno}
\usepackage{tabularx}
\usepackage{changes}
\usepackage{booktabs}

\title{VECTOR: Velocity-Enhanced GRU Neural Network for Real-Time 3D UAV Trajectory Prediction}

\author{%
\href{https://orcid.org/0000-0001-7493-9318}{\includegraphics[scale=0.06]{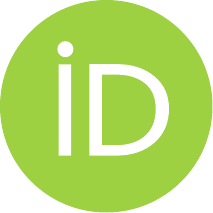}\hspace{1mm}{\textbf{Omer Nacar}}}\thanks{Corresponding author: \href{mailto:onajar@psu.edu.sa}{onajar@psu.edu.sa}}\\ \and
\href{https://orcid.org/0000-0002-0518-852X}{\includegraphics[scale=0.06]{orcid.pdf}\hspace{1mm}{\textbf{Mohamed Abdelkader}}}\\ \and
\href{https://orcid.org/0000-0002-6381-4250}{\includegraphics[scale=0.06]{orcid.pdf}\hspace{1mm}{\textbf{Lahouari Ghouti}}}\\ \and
\textbf{Khaled Hammad}\\ \and
\href{https://orcid.org/0009-0007-1804-1109}{\includegraphics[scale=0.06]{orcid.pdf}\hspace{1mm}{\textbf{Abdulrahman Al-Batati}}}\\ \and
\href{https://orcid.org/0000-0003-3787-7423}{\includegraphics[scale=0.06]{orcid.pdf}\hspace{1mm}{\textbf{Anis Koubaa}}}}

\date{%
\vspace{1em}%
All authors are affiliated with the College of Computer and Information Sciences,\\
Prince Sultan University, Riyadh, Saudi Arabia\\[1ex]
\texttt{\{onajar, mabdelkader, lghouti, khammad, aalbatati, akoubaa\}@psu.edu.sa}}

\renewcommand{\headeright}{}
\renewcommand{\undertitle}{}
\renewcommand{\shorttitle}{VECTOR: Velocity-Enhanced GRU Neural Network for Real-Time 3D UAV Trajectory Prediction}

\begin{document}
\maketitle

\begin{abstract}
	This paper addresses the challenge of predicting 3D trajectories for Unmanned Aerial Vehicles (UAVs) in real-time, a critical task for applications like aerial surveillance and defense. Current prediction models primarily leverage only position data, which may not provide the most accurate forecasts for UAV movements and usually fail outside the position domain used in the training phase. Our research identifies a gap in utilizing velocity estimates, first-order dynamics, to better capture the dynamics and enhance prediction accuracy and generalizability in any position domain.
To bridge this gap, we introduce a trajectory prediction scheme, using sequence-based neural networks with Gated Recurrent Units (GRUs), to forecast future velocity and positions based on velocity historical estimates instead of position measurements. This approach is designed to improve the predictive capabilities over traditional methods that rely solely on recurrent neural networks (RNNs) or transformers, which can struggle with scalability in this context.
Our methodology employs both synthetic and real-world 3D UAV trajectory data, incorporating diverse patterns of agility, curvature, and speed. Synthetic data is generated using the Gazebo robotics simulator and PX4 Autopilot, while real-world data is sourced from the UZH-FPV and Mid-Air drone racing datasets. We train the GRU-based models on drone 3D position and velocity samples to capture the dynamics of UAV movements effectively.
Quantitatively, the proposed GRU-based prediction algorithm demonstrates superior performance, achieving a mean square error (MSE) ranging from $2 \times 10^{-8}$ to $2 \times 10^{-7}$. This performance outstrips existing state-of-the-art RNN models. Overall, our findings confirm the effectiveness of incorporating velocity data in improving the accuracy of UAV trajectory predictions across both synthetic and real-world scenarios, in and out of position data distributions. Finally, we open-source our 5000 trajectories dataset and a ROS 2 package to facilitate the integration with existing ROS-based UAV systems.
\end{abstract}


\section*{Supplementary Material}
\begin{itemize}
    \item A video of the simulation and real experiments is availabe at \url{https://youtu.be/CDp69R_izqo}
    \item Open-source code is available at \url{https://github.com/riotu-lab/GRUTrajectoryPredictor.git}
    \item a ROS2 package is available at \url{https://github.com/mzahana/drone_path_predictor_ros}
\end{itemize}

\section{INTRODUCTION}\label{sec:intro}

Trajectory prediction has become a pivotal component in the realm of autonomous systems, encompassing a wide range of applications such as pedestrian movement in urban environments, dynamic flight paths of Unmanned Aerial Vehicles (UAVs), and the navigation of autonomous cars \cite{shrivastava2021deep,zhong2022short,jiang2019trajectory}. Accurately predicting trajectories is fundamental to ensuring safety, efficiency, and reliability in these systems. For pedestrians, precise trajectory prediction can improve safety and flow in crowded spaces. In the context of UAVs, it is crucial for avoiding collisions, managing airspace, and achieving efficient route planning. Similarly, for autonomous vehicles, predicting the movement of surrounding objects is essential for safe navigation, traffic management, remote sensing observation, vehicle detection, and UAV detection \cite{perez2018path,tan2021trajectory, rs15071873,electronics10070820,9750351}. Unlike the prediction of pedestrians motion which is much less agile compared to UAVs, UAV trajectory prediction complexity is compounded by the necessity to operate in real-time and adapt to varying UAV behaviors.

\subsection{Problem Statement}\label{sec:problem}

Despite significant advancements in UAV trajectory prediction techniques, predicting trajectories in 3D environments remains a challenging task due to the UAVs' high agility and non-linear dynamics. This complexity is further amplified in real-time scenarios, where rapid and accurate predictions are crucial. While many existing methods rely on historical data, they often fall short in effectively capturing the complex, dynamic variability of UAV trajectories, particularly when using only positional information or simplistic models that cannot fully generalize across different trajectory types.

Our approach addresses these limitations by simply incorporating first-order dynamics that can be captured by velocity measurements, enabling the model to better capture the intricate dynamics of UAV movements. By leveraging sequence-based neural networks equipped with Gated Recurrent Units (GRUs), our prediction model efficiently encodes the UAV's trajectories history and enhances memory capacity. The model's performance is demonstrated on both synthetic and real-world datasets, proving its ability to accurately predict UAV trajectories under diverse conditions. Furthermore, the versatility of our model allows for seamless extension to other autonomous systems through retraining with relevant datasets, offering a robust solution to the challenges of real-time trajectory prediction.
\subsection{Related Works} \label{sec:related_works}

Trajectory prediction is considered critical in many robotics and autonomous systems applications, involving forecasting future paths of moving units based on historical data. Applications, including pedestrian movement prediction in urban environments and autonomous vehicle/UAV navigation, have evolved significantly over the recent years. While pedestrian and vehicle trajectory prediction primarily focuses on 2D spatial data, UAV-focused trajectory prediction problems extend are formulated in 3D spaces which adds another complexity dimension. Given the additional complexity, efficient preprocessing and prediction solutions should be devised especially when dealing with real-time UAV flight movements.

Considerable advancements have been made in the problem of pedestrian trajectory prediction. Early approaches, rule-based, employ human-crafted rules and symbolic pattern extraction \cite{shrivastava2021deep,monreale2009wherenext}. Newer ones resort to unsupervised machine learning (ML) techniques such as {\em clustering} \cite{morzy2007mining,roh2010nncluster} to predict future routes based on trajectory similarity. Latest solutions incorporate neural network (CNN) models \cite{zamboni2022pedestrian}. CNN-based models outperform rule- and clustering-based ones. This substantial prediction performance boost has paved the way for more solutions relying on deep learning (DL) algorithms and reflects ongoing efforts to continually improve the precision and reliability of prediction solutions to handle complex and dynamic 2D environments.

Since vehicle path prediction ensures the safety and efficiency of autonomous driving systems, substantial contributions to this problem have been reported in the literature
\cite{jiang2019trajectory,won2009trajectory, ashbrook2003using,yao2018deep}. Initial solutions considered hybrid approaches that combine clustering and hidden Markov models (HMMs) \cite{won2009trajectory,ashbrook2003using} marking the enthusiastic adoption of conventional ML frameworks for vehicle trajectory prediction. Larger DL sequence-based neural architectures such as long-short term memory (LSTM) and GRUs contributed to the prediction accuracy improvement \cite{jiang2019trajectory,yao2018deep}. In \cite{jiang2019trajectory}, Jiang et al. proposed an RNN-based prediction scheme that outperformed other architectures for the prediction of autonomous vehicles positions and velocities\cite{jiang2019trajectory}. The reported results corroborate the ability of RNN models to properly model mobility data history though constant update of the intrinsic model memory. In addition, the advancement in vehicle trajectory prediction models underscores the effectiveness achieved to address the challenges of dynamic and unpredictable traffic environments.

As stated above, the 3D aspect of UAV trajectory prediction adds further complexity to the underlying problem. Unlike 2D settings,  UAVs operate in 3D spaces and require sophisticated modeling to anticipate  more complex trajectory patterns. In addition to enhancing the security and efficacy of UAV operations, UAV trajectory prediction opens up new possibilities for innovative applications in various disciplines including environmental monitoring, delivery, and surveillance. UAV trajectory prediction has made substantial progress where novel methods and various uses are identified to address UAV operations across multiple sectors. 

Yang et al. proposed a bidirectional (BiGRU) model to predict the trajectories of small-scale quadrotors \cite{yang2020real}. Compared to traditional GRU configurations, the BiGRU-based model has demonstrated superior performance in all scenarios in terms of faster convergence and improved prediction accuracy. The reported performance scores justify the potential of bidirectional neural architectures to capture the underlying dynamics during the real-time prediction of UAV trajectories. Also, the BiGRU-based solution easily adapts to different quadrotor types by relying only on historical location data.

Many applications, related to UAV trajectory prediction, are reported in the literature. For instance, Zhang et al. leveraged automatic dependent surveillance broadcast (ADS-B) technology using a recurrent LSTM (RLSTM) technique \cite{zhang2022recurrent} to predict UAV trajectories. The technique attained satisfactory average trajectory prediction error scores. The integration of ADS-B for UAV positioning, combined with the RLSTM model, represents a novel approach to trajectory prediction, especially for air traffic control and collision avoidance. In \cite{zhu2022short}, Zhu et al. presented a GRU-based scheme for short-term trajectory prediction of drones. Their scheme is reported to be particularly effective in low-altitude airspace. Using historical trajectory data, Zhu et al. assessed the model performance based on one-step and multi-step predictions. 

A novel application, attributed to Tan et al. \cite{tan2021trajectory}, introduces control and software architectures using ROS C++ for UAVs to intercept objects with on-board depth cameras. Unlike previous trajectory prediction methods, relying on off-board tools, Tan et al. rely on the on-board capability for object interception using iterative trajectory prediction and point cloud processing. Gazebo-based simulations results emphasize the potential of this architecture in enhancing the on-board autonomy of UAVs.

The literature shows significant progress in UAV trajectory prediction, with modern models often relying on bidirectional deep learning architectures and ADS-B technology. These models focus on improving accuracy, adaptability, and real-time prediction to meet the growing demands of UAV navigation in diverse environments. However, many existing approaches still face challenges in fully capturing the dynamic nature of UAV movements with fast changing maneuvers and high speeds. Our work contributes to this field by addressing these limitations. Specifically, we use velocity estimates derived from position information to create a more comprehensive understanding of UAV dynamics. This integration fills a key gap in the current research, where velocity is often underutilized. Additionally, we demonstrate the accuracy and generalizability of real-time position trajectory prediction based on velocity prediction models in realistic simulation and real-world experiments, showing our model robustness against unseen position messmates.

\subsection{Contributions}

This paper presents a comprehensive solution for UAV trajectory prediction in 3D environments, comparing position and velocity historical data with sequence-based neural networks of varying complexity. Our key contributions are summarized as follows:

\begin{itemize}
    \item \textbf{Development of Real-time 3D Trajectory Prediction Model:} We introduce a real-time UAV trajectory prediction model based on sequence-based neural networks utilizing GRU cells. This model, with 24 different configurations, efficiently handles the dynamic and unpredictable nature of UAV movements and achieves real-time inference rates of up to 100Hz on edge devices like the Jetson Orin NX.
    
    \item \textbf{Creation and Utilization of Diverse 3D UAV Trajectory Datasets:} We build a comprehensive dataset of synthetic 3D UAV trajectories using the {\em Gazebo robotics simulator} and {\em PX4 autopilot}, augmented with publicly available datasets. This dataset enables rigorous training and evaluation of our models and supports reproducible research.

    \item \textbf{Real-world Experiments and Sim2Real Validation:} We validate the effectiveness of our models through real-world experiments. Models trained exclusively on simulated trajectories (ROS-Gazebo) were successfully applied in real-world tests, demonstrating the robustness of the sim2real concept and the practical applicability of our approach.

    \item \textbf{Extensive Performance Evaluation:} We conduct a detailed analysis of 24 model configurations, validating our approach with both synthetic and real-world UAV trajectories. This evaluation confirms the superior accuracy and robustness of our prediction models compared to existing methods.
\end{itemize}

Together, these contributions mark a significant advancement in the field of UAV trajectory prediction, offering a practical and scalable solution for real-time applications.

\subsection{Paper Outline}
To lay a proper foundation for our research work, the paper is organized as follows. Section \ref{sec:intro} introduces the problem of UAV trajectory prediction in 3D settings where the rational behind the use of sequence-based models is outlined. The main challenges and issues, associated with the prediction of UAV trajectories in highly dynamic environments, are discussed therein. Then, Section \ref{sec:dataset} details the adopted process for the creation of the synthetic datasets of drone trajectories in 3D settings using {\em Gazebo robotics simulator} and {\em PX4 Autopilot}. In addition, this section provides valid justifications for the selected simulation settings to account for drone trajectories with varying agility patterns, curvatures and speeds. A brief review of the selected real-world drone trajectories, {\em UZH-FPV} and {\em Mid-Air} drone racing datasets, is given. The detailed description of the proposed sequence-based prediction model can be found in Section \ref{sec:model_architecture} where we discuss the design choices that form the foundations of the sequence-based architectures and their varying configurations to enable efficient and longer memory capacity. In Section \ref{sec:experiment}), the experiment design to train and evaluate the proposed sequence-based prediction model is laid out where we discuss the selected performance metrics to assess the performance of the proposed and existing models. The results of the extensive performance evaluation experiments are summarized in Section \ref{sec:results} where a detailed discussion is provided to highlight the main reasons behind the superiority of our sequence-based model over existing models in terms of prediction accuracy and  generalization capability. Finally, Section \ref{sec:conclusions} concludes this paper where we summarize the key findings, contributions and future extensions of our research.

\section{UAV 3D Trajectory Datasets}
\label{sec:dataset}
In this work, the datasets consist of a range of short and long drone trajectories along with real-time simulation datasets, capturing the diverse dynamics of UAV flight in 3D space settings. We use a multi-faceted method to make the most of these datasets, including careful preprocessing, normalization, and statistical analysis to guarantee that the data we feed into our models is clean, representative, and supportive of reliable machine learning. The datasets used include both publicly available data\footnote{https://huggingface.co/datasets/riotu-lab/Synthetic-UAV-Flight-Trajectories}, including FPV-UZH \cite{delmerico2019we} and MID-AIR \cite{Fonder2019MidAir} datasets with our custom simulations, providing a comprehensive view of UAV flight patterns in different environments.

\subsection{Overview of the UAV Trajectory Datasets}
\begin{figure*}[ht]
\centering
\begin{subfigure}{.32\textwidth}
  \centering
  \includegraphics[width=.95\linewidth]{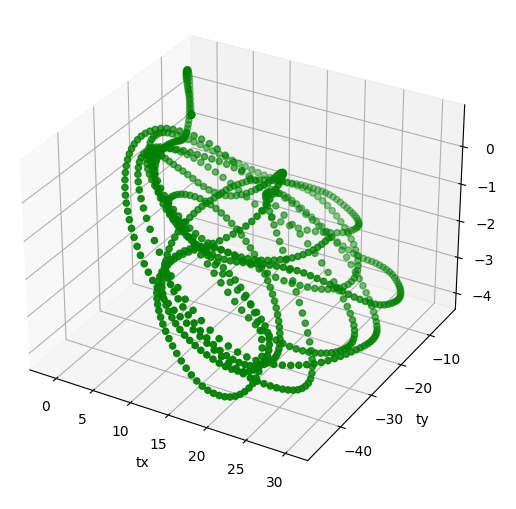}
  \caption{Short UAV traj. sample}
  \label{fig:random_uzh_traj}
\end{subfigure}%
\begin{subfigure}{.32\textwidth}
  \centering
  \includegraphics[width=.95\linewidth]{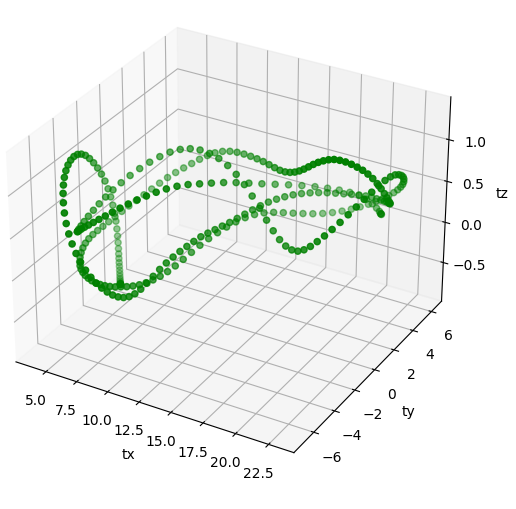}
  \caption{Short UAV traj. sample}
  \label{fig:random_uzh_traj_2}
\end{subfigure}%
\begin{subfigure}{.32\textwidth}
  \centering
  \includegraphics[width=.95\linewidth]{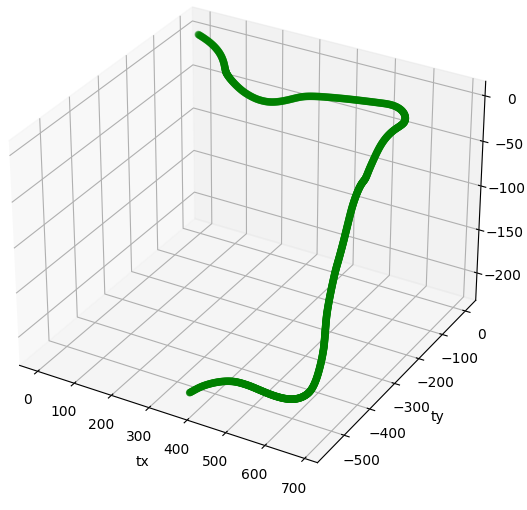}
  \caption{Long UAV traj. sample}
  \label{fig:random_midair_traj}
\end{subfigure}

\begin{subfigure}{.32\textwidth}
  \centering
  \includegraphics[width=.95\linewidth]{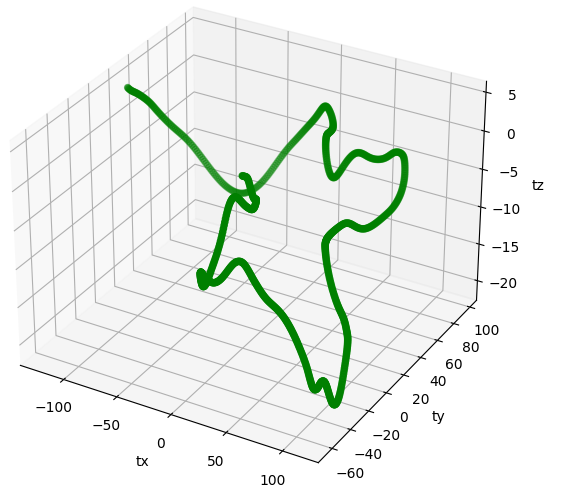}
  \caption{Long UAV traj. sample}
  \label{fig:random_midair_traj_2}
\end{subfigure}%
\begin{subfigure}{.32\textwidth}
  \centering
  \includegraphics[width=.95\linewidth]{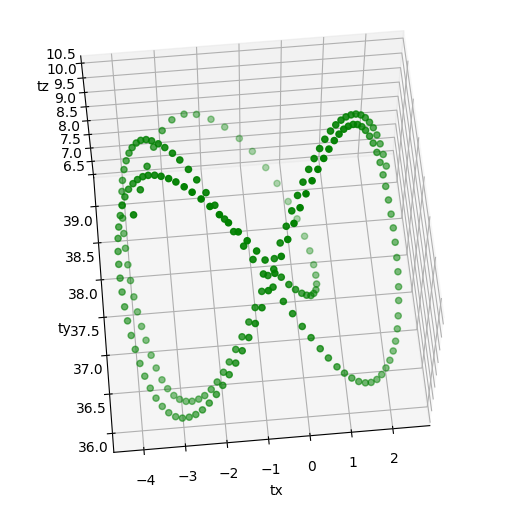}
  \caption{Infinity simulated UAV traj.}
  \label{fig:random_sim_traj_inf}
\end{subfigure}%
\begin{subfigure}{.32\textwidth}
  \centering
  \includegraphics[width=.95\linewidth]{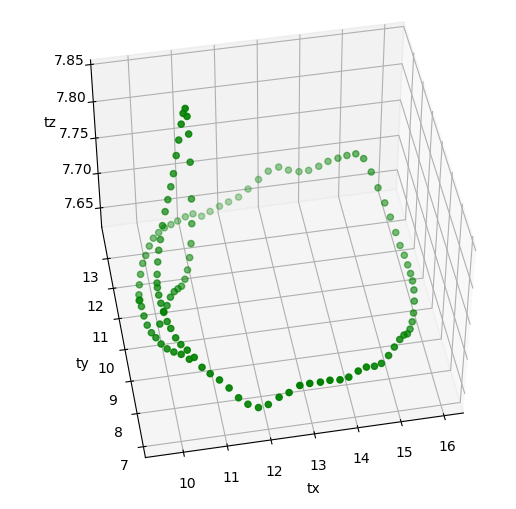}
  \caption{Circle simulated UAV traj.}
  \label{fig:andom_sim_traj_2}
\end{subfigure}

\caption{Composition of Various Raw Data Sources Used for GRU Model Training.}
\label{fig:raw_dataset}
\end{figure*}

A diverse set of datasets is used to train and evaluate the GRU models for drone trajectory forecasting. The trained dataset includes two publicly available datasets, FPV-UZH and MID-AIR, and we also added a large, simulated dataset that we have generated to generalize the model to real-time drone trajectory prediction scenarios. The FPV-UZH  targets short trajectory data, primarily focusing on fast-paced, dynamic drone movements. It is characterized by its high-resolution spatial data and temporal granularity. In contrast, the Mid-Air is considered for its long trajectory data, which includes extensive flight paths, capturing various environmental conditions, and longer-duration flights.
Moreover, we generated simulated drone trajectories using the Gazebo simulation environment to generalize our models for better real-time simulation scenarios. This dataset is tailored to cover various scenarios, including line, circle, and infinity-shaped drone trajectory paths, providing more complex movements not fully represented in the FPV-UZH and Mid-Air datasets. These three datasets were mixed to form a comprehensive training dataset, which resulted in more than 500K inputs and output segments with a timestamp step size of 0.1 and a sequence length of 20 and 10, respectively. Figure \ref{fig:raw_dataset} demonstrates the various raw datasets combined to create our comprehensive dataset for processing and training the GRU models.

\subsection{Data Preprocessing and Preparation}
We proposed an essential modification of data processing in our experimental setup by calculating velocity from the positional data in our mixed and simulation-only datasets. This method is motivated by the concept that velocity might provide a more consistent and general feature for drone trajectory prediction. Unlike position, which may vary significantly based on the drone's operational context, specific spatial coordinates do not limit velocity. Instead, it offers a consistent and continuous illustration of movement, which might enable the GRU model to more accurately generalize the prediction of the next drone's movements regardless of the position. We used both position and velocity data to explore this hypothesis and to train the GRU models separately. This comparative analysis aims to assess how well each type of data performs in terms of prediction accuracy, generalization to unseen test datasets, and model adaptability. According to our theory, velocity data, relative consistency, and range may contribute to a more comprehensive knowledge of drone dynamics, even while positional data provide explicit spatial information.

Moreover, normalization plays a crucial role in improving model performance. It ensures that all input features contribute equally to the learning process, reducing the dominance of certain features and minimizing the risk of poor convergence. Without proper normalization, features with larger values (e.g., position in a 3D space) could overshadow others (e.g., velocity), which might lead to suboptimal predictions, particularly in models dealing with dynamic and nonlinear data like UAV trajectories. Furthermore, normalizing helps prevent issues arising from correlated features, which could hinder the model’s ability to capture the true underlying patterns in the data. 

In this work, We first apply a Whitening normalization process to the simulated data by computing the Cholesky decomposition of the data covariance matrix. The normalization factors used in the pre-processing step are summarized in Table \ref{tab:Normstat}. Given the covariance matrix, $ \Sigma $, the Cholesky decomposition is represented using $\Sigma = L L^T$, where $ L $ is a lower triangular matrix \cite{huang2019iterative}. Data whitening aims to mitigate the feature correlation effects. For instance, whitening a position vector, $ \mathbf{p} $, is achieved as shown in equation \eqref{eq1}

\begin{equation}\label{eq1}
\mathbf{p}_{w} = L^{-1}(\mathbf{p} - \mu)
\end{equation}
where $ \mathbf{p} $ and $ \mathbf{p}_{w} $ represent the normal and whitened position vectors. The mean position vector is given by $ \mu $. $ L^{-1} $ represents the inverse matrix of $ L $.

Conversely, dewhitening involves reverting the whitened data to its original state, which is expressed by equation \eqref{eq2}:
\begin{equation}\label{eq2}
\mathbf{p} = L \cdot \mathbf{p}_{w} + \mu
\end{equation}

Additionally, since UAV trajectories often involve nonlinear dynamics with strong couplings between position and velocity components, we also propose another normalization method to further preserve the integrity of these dynamics. Direct transformations could alter the direction of the vectors, which would misrepresent the original UAV motion. To prevent this, we normalize position and velocity vectors against the vector with the maximum $L_2$-norm \cite{patro2015normalization}, thereby maintaining the essential properties of the data. The $L_2$-norm normalization ensures that the scaling of the vectors does not distort the original dynamic behavior. The maximum $L_2$-norm is defined in equation~\ref{eq3}:

\begin{equation}\label{eq3}
\lVert \mathbf{p} \rVert_{\text{max}} = \max_{i} \sqrt{x_{i}^{2} + y_{i}^{2} + z_{i}^{2}}
\end{equation}

This method prevents extreme values from skewing the normalization process, ensuring that the dynamic relationships between position and velocity remain intact, even after scaling.

\begin{table}[ht]
\caption{Pre-processing Statistics for Position and Velocity UAV Dataset}
\centering
\begin{tabular}{|c|c|c|}
\hline
\textbf{Stats} & \textbf{Position} & \textbf{Velocity} \\ \hline
Mean & [0.423, 0.501, 11.764] & [0.004, 0.001, 0.003] \\ \hline
Std & [23.072, 22.913, 4.436] & [1.817, 1.706, 0.792] \\ \hline
Cov Mat & 
\begin{tabular}{@{}c@{}}
532.298, -26.305, -3.141, \\
-26.305, 525.017, 1.172, \\
-3.141, 1.172, 19.678
\end{tabular} & 
\begin{tabular}{@{}c@{}}
3.301, -0.023, 0.002, \\
-0.023, 2.910, 0.047, \\
0.002, 0.047, 0.628
\end{tabular} \\ \hline
L Mat & 
\begin{tabular}{@{}c@{}}
0.043, 0.0, 0.0 \\
0.002, 0.044, 0.0, \\
0.007, -0.003, 0.225
\end{tabular} & 
\begin{tabular}{@{}c@{}}
0.550, 0.0, 0.0, \\
0.004, 0.587, 0.0, \\
-0.002, -0.044, 1.262
\end{tabular} \\ \hline
Max Len/Vel & 61.765 & 8.002 \\ \hline
\end{tabular}
\label{tab:Normstat}
\end{table}

Each normalization technique (whitening and $L_{2}$-norm nornmalization) adds a distinct benefit to the model's capacity to learn and predict drone trajectories, and together, they maximize the data used to train our GRU models.

\subsection{Automated Generation of UAV Datasets}

The paper proposes an automatic generation of datasets for Unmanned Aerial Vehicles (UAVs) that follow predetermined and randomly generated trajectories. We use time-parameterized trajectories, which allow trajectory points to be generated in real time while running simulations. To be more precise, we present two categories of trajectories: patterns that resemble infinity and circles. Because of the flexibility of our technique, several parametric trajectory models can be integrated. Throughout our simulation, the UAV concurrently gathers a wide range of sensor data and follows its predetermined path. A key feature of our system is the temporal synchronization of these data points, ensuring their utility for subsequent analytical processes.

\subsubsection{Circular Trajectory in 3D Space}
The circular trajectory in 3D space is characterized by its center, radius, and orientation in space. The trajectory at any given time is determined as follows:

\begin{itemize}
    \item \textbf{Normal Vector:} $\mathbf{n} = \frac{\mathbf{n}_{orig}}{\lVert \mathbf{n}_{orig} \rVert}$, where $ \mathbf{n}_{orig} $ is an arbitrary average vector that is normal to the plan where the circular trajectory is defined. The norm of this vector is given by $ \lVert \mathbf{n}_{orig} \rVert $.
    \item \textbf{Center Vector:} $\mathbf{c}$, representing the center of the circle.
    \item \textbf{Radius:} $r$, the radius of the circle.
    \item \textbf{Angular Velocity:} $\omega$, defining the rate of traversal along the trajectory.
\end{itemize}

The position on the trajectory at time $t$, $\mathbf{p}_c(t) \in \mathbb{R}^3$ is given by equation \eqref{eq4}:
\begin{equation}\label{eq4}
    \mathbf{p}_c(t) = \mathbf{c} + r \cdot (\cos(\omega t) \cdot \mathbf{v_1} + \sin(\omega t) \cdot \mathbf{v_2}),
\end{equation}
where $\mathbf{v_1}$ and $\mathbf{v_2}$ are orthogonal unit vectors in the plane of the circle, derived from the normal vector $\mathbf{n}$.

\subsubsection{Infinity-Like Trajectory in 3D Space }
The infinity-like trajectory shares the same initial parameters as the circular trajectory but differs in the trajectory equation. The normal vector, center vector, radius, and angular velocity are defined as in the circular trajectory. The position on the trajectory at time $t$, $\mathbf{p}_\infty (t) \in \mathbb{R}^3$ is defined as equation \eqref{eq5}:

\begin{equation}\label{eq5}
    \mathbf{p}_{\infty}(t) = \mathbf{c} + r \cdot (\cos(\omega t) \cdot \mathbf{v_1} + \sin(2\omega t) \cdot \mathbf{v_2}),
\end{equation}

where $\mathbf{v_1}$ and $\mathbf{v_2}$ are defined as in the circular trajectory. The factor of $2$ in the $\sin$ function's argument for the $\mathbf{v_2}$ component creates the infinity-like pattern. Table \ref{tab:ParameterBounds} shows the parameter bounds used for the generation of circle and infinity trajectories:

\begin{table}[ht]
\centering
\caption{Parameter Bounds for Circle and Infinity Trajectories}
\label{tab:traj_parameter_bounds}
\begin{tabular}{|l|c|}
\hline
\textbf{Parameter}    & \textbf{Bounds} \\
\hline
Center, $\mathbf{c}$ (m)               & \((-40.0, 40.0, 5)\) to \((40.0, 40.0, 20.0)\) \\
\hline
Normal Vector, $\mathbf{n}$ (m)         & \((-40.0, 40.0, 5)\) to \((40.0, 40.0, 20.0)\) \\
\hline
Radius, $r$ (m)               & \([1.0, 5.0]\) \\
\hline
Speed, $\omega$ (rad/s)                 & \([0.3, 1.0]\) \\
\hline
\end{tabular}
\label{tab:ParameterBounds}
\end{table}

Based on the parameters and bounds provided in Table \ref{tab:ParameterBounds}, two types of parametric trajectories of circular and infinity-like shapes are generated, ensuring that an arbitrary number of these trajectories can be generated by altering their parameters.
\section{Model Architecture}\label{sec:model_architecture}

In this work, we investigate using a Gated Recurrent Unit (GRU) model with different structural configurations and GRU encoding and decoding layers to examine the relationship between model complexity and drone trajectory prediction accuracy. The GRU, a derivative of the Recurrent Neural Network (RNN) family, integrates gating techniques essential to controlling information flow between neural network cells. The GRU architecture was first presented by \cite{cho2014learning}, and it is specifically made to capture dependencies in extensive sequential data without losing essential information from previous segments of the sequence. This capability is primarily facilitated by its update and reset gating techniques, similar to those that exist in Long Short-Term Memory (LSTM) networks, which effectively address the vanishing and exploding gradient issues that traditionally occur in RNNs \cite{pascanu2012understanding}. These gates are crucial in determining the information to be kept or discarded at each time step. 

At its core, the GRU processes sequential input data iteratively. During each time step, the GRU cell receives a batch of input data. The collected batch data covers a limited time span such as $ 20 $ seconds. The collected time span data includes drone trajectory coordinates recorded over $ 20 $ seconds. Thanks to its internal memory structure, GRU cells can keep track of the drone dynamics over the given time span using its hidden state commonly known as $ h_{t-1} $. This hidden state, encapsulating the information from prior data points in the sequence, is subsequently merged with the incoming input segment for the next step. This cyclic procedure ensures a continuous update and integration of information throughout the sequence, contributing to the prediction of the next drone positions and velocities over a time span of $ 10 $ seconds. Therefore, fast predictions over limited time horizon can be carried out without burdening the GRU memory. Relying on wider time spans may result in poor GRU memory due to vanishing/exploding gradients. Therefore, our  methodology allows the GRU's proficiency to learn and predict the dynamic changes in the UAV's trajectory without risking a degradation in prediction accuracy.

\begin{figure}[t]
    \centering
    \includegraphics[width=0.7\textwidth]{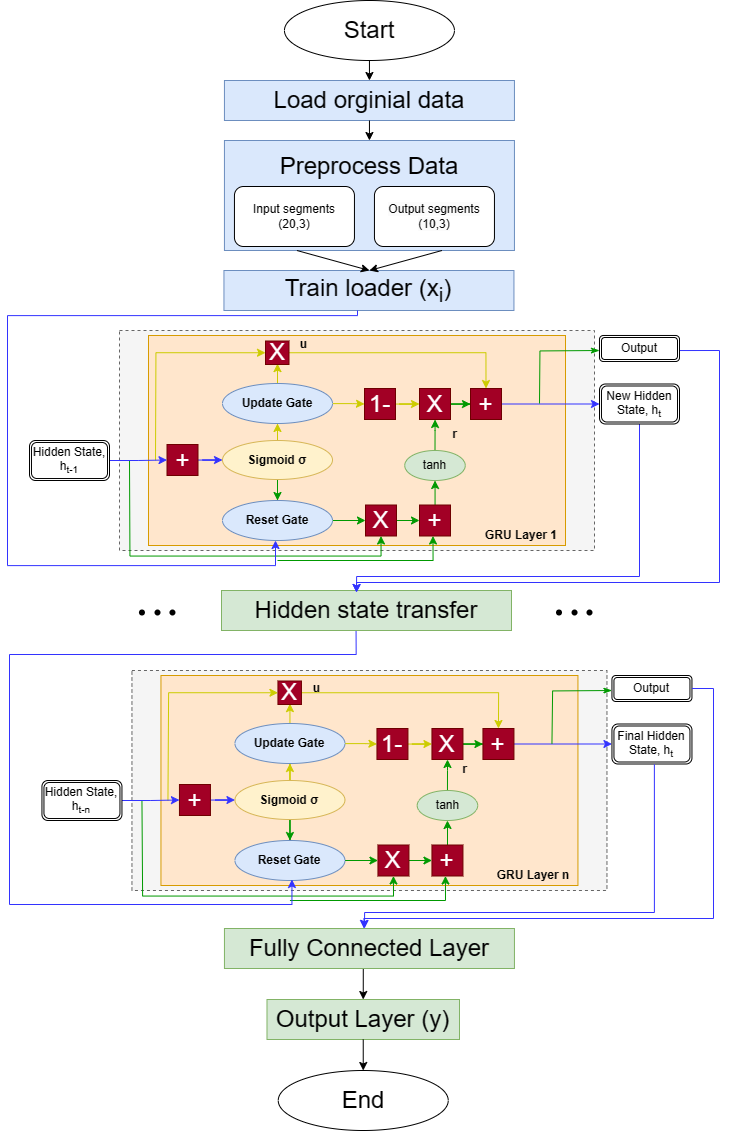}
    \caption{The flowchart of the GRU algorithm for UAV 3D trajectory prediction}
    \label{fig:Gruflowchart}
\end{figure}

Figure \ref{fig:Gruflowchart} demonstrates the flowchart of the GRU algorithm for UAV 3D trajectory prediction along with the inner workings of the GRU layers and gates.

The hidden state of the GRU, denoted as $h_{t-1}$, plays a pivotal role in capturing the historical context of the UAV's movement. It is a memory unit, retaining essential information from previous trajectory points. As new trajectory data (a new segment of coordinates) is fed into the GRU, it combines with the existing hidden state to update the model's understanding of the UAV's movement pattern. As depicted in Figure \ref{fig:Gruflowchart}, the update and reset gates within the GRU architecture are particularly beneficial for trajectory prediction. These gates effectively manage the information flow, ensuring the model retains relevant data (patterns or trends) and filters out irrelevant details (such as noise). This selective memory mechanism is crucial for accurately predicting future positions or velocities of the UAV, as it allows the model to focus on significant trajectory patterns while ignoring minor fluctuations. The GRU model's computations produce a new hidden state ($h_t$) at each time step, informed by both the current input and the previous hidden state. This updated hidden state is then used to predict the next segment of the UAV's trajectory.

The inner structure of the GRU layers can be broken down into three main mechanisms: 1) reset; 2) update and 3) hidden state steps. In the reset step, the output is computed by combining the previous hidden state ($ h_{t-1} $) and the current input using element-wise multiplication and summing prior to non-linear activation through a sigmoid function. The non-linear activation sets the the output values between  0 and 1. Such process allows the gate to filter out the least-important information while keeping the most-important one that will be further processed in the subsequent steps. The reset gate process is expressed using equation \eqref{eq6}

\begin{equation}
\text{gate}_{\text{reset}} = \sigma(W_{\text{input reset}} \cdot x_t + W_{\text{hidden reset}} \cdot h_{t-1})
\label{eq6}
\end{equation}

where, $ W_{input_{rest}} $ and $ W_{hidden_{rest}} $ are the trainable weight matrices associated with the previous hidden state and the current input, respectively. $ x_{t} $ represent the input and $ h_{t-1}$ is the previous hidden state. 

The GRU model is trained through propagation through many iterations while updating the weight metrics to retain only valuable features. This is done using the previous hidden state, which will first be multiplied by a trainable weight and then undergo an element-wise multiplication (Hadamard product) with the reset output. At the same time, the current input will also be multiplied by a trainable weight before being summed with the product of the reset vector and the previous hidden state above. Lastly, a non-linear activation \emph{tanh} function will be applied to the final output $ r $ in equation \eqref{eq7}
\begin{equation}
r = \tanh(\text{gate}_{\text{reset}} \odot (W_{h1} \cdot h_{t-1}) + W_{x1} \cdot x_t)
\label{eq7}
\end{equation}

The update gate is computed using the previous hidden state and current input data in a similar fashion to the reset gate, but the weights multiplied with the input and hidden state are unique to each gate, which means that the final vectors for each gate are different as shown in equation \eqref{eq8}

\begin{equation}
\text{gate}_{\text{update}} = \sigma(W_{\text{input update}} \cdot x_t + W_{\text{hidden update}} \cdot h_{t-1})
\label{eq8}
\end{equation}

The updated output will then undergo element-wise multiplication with the previous hidden state to obtain $u$ in equation \eqref{eq9}, which will be used to compute our final output later:

\begin{equation}
u = \text{gate}_{\text{update}} \odot h_{t-1}
\label{eq9}
\end{equation}

The quantity $ u $, calculated in Equation \ref{eq6}, is used later to get the final output. Moreover, with this algorithm, the update gate can serve well in helping the model determine how much of the past information stored in the previous hidden state needs to be retained for the future. In order to obtain the updated hidden state, the update gate will be reused by taking the element-wise inverse version of the same update vector (1 - update gate) and doing an element-wise multiplication with our output from the Reset gate, $ r $. That results in determining what part of the new information should be stored in the hidden state. Finally, by summing the result from the operation above with the output of the update gate in the previous step (i.e., $ u $). This will result in the updated hidden state as defined in equation \eqref{eq10}

\begin{equation}
h_t = r \odot (1 - \text{gate}_{\text{update}}) + u
\label{eq10}
\end{equation}

where $ h_{t} $ is the hidden state, $r$ is the output of the reset gate, $\odot$ denotes the Hadamard product (element-wise multiplication), and $u$ is the output of the product of the update gate and the previous hidden state, representing the preserved information from the past.

This overall implementation of the Gated Recurrent Unit (GRU) model is used in UAV 3D trajectory prediction, which underscores its efficacy in handling complex sequential data of the positions and velocities of the trajectories. The GRU's architecture, characterized by its update and reset gates, adeptly manages information flow, thereby ensuring the preservation of relevant trajectory patterns while filtering out noise. 

\section{Experiment}\label{sec:experiment}

The experimental setup used in this paper is to assess how well Gated Recurrent Unit (GRU) models predict drone trajectories for position and velocity segments. Our experimental framework provides a comprehensive dataset preparation, model architecture, training procedures, and evaluation metrics, focusing on the model's performance under various conditions, including model complexity and normalization steps.

\subsection{Dataset and Preprocessing}

This experiment's training dataset comprises a mixed dataset of short and long historical 3D UAV trajectories and real-time drone trajectories. The real-time simulation dataset consists of 5375 UAV position trajectories collected from 20-hour simulated flights. We also used a simulation-only dataset and investigated the behavior of the GRU model on it separately. 

\subsubsection*{Velocity Segments from Position Data}

We transformed position segments into corresponding velocity segments to enhance the depth and diversity of our dataset. This process involved calculating velocity segments using a first-order approximation based on the rate of change in position trajectories. The velocity trajectories were computed by approximating the change in position over time. Then, the trajectories were resampled at $T_{s}= 0.1$ second intervals. Subsequently, these were segmented into distinct input and output segments for modeling purposes. The dataset includes more than 655,728 input and output segments for position and 650,353 for velocity, with a sampling time of 0.1 seconds. Each input segment spans 2.0 seconds, resulting in 20 data points, and each output segment length covers 1.0 seconds, corresponding to 10 data points. Furthermore, we used two different normalization techniques involving normalizing the position or velocity vectors by using the vector that exhibits the maximum 2-norm (maximum distance or speed) and the whitening method described in Section \ref{sec:dataset}.

This methodology ensures a comprehensive representation of UAV movement and aligns with the need for precise spatial and temporal dimensions modeling in UAV trajectory prediction.

\subsection{Model Configuration and Structure}

The model architecture and configuration used in this experiment are primarily focused on investigating the model complexity and how it might affect the GRU-based Trajectory Predictor's performance. The architecture of the GRU model is pivotal in understanding different patterns and trends of drone trajectory prediction. The Trajectory Predictor, as shown in Figure \ref{fig:Gruflowchart}, implemented in PyTorch, comprises an encoder-decoder structure using GRU layers and a fully connected output layer, providing a robust framework for capturing the temporal dynamics of trajectories. The model's encoder uses a GRU layer incorporating dropout regularization to reduce overfitting. The decoder, which is an additional GRU layer, is reconstructing the trajectory output. A fully connected layer converts The decoder output into the final projected trajectory. Thanks to its architecture, the model can efficiently learn complex temporal patterns from drone flying data. 

The structure of the GRU model has involved a variety of the model's complexity by adjusting the hidden dimension and layer count configurations, keeping the input (3D coordinates) and output dimensions constant. In particular, we looked into three setups, including the default design consisting of 64 hidden units with two layers, which provides a good compromise between computational performance and model complexity. Then, we considered 128 three-layered hidden units with more intricate models, which might anticipate extracting more information and flight trends and potentially enhancing predictions but come at the expense of additional processing power. After that, a model with five layers and 256 hidden units is the most complicated model in our analysis. It was created to see how well our architecture could handle the mixed and simulation-only dataset. 

\subsection{Model Training}

In the training phase, the mixed and simulation-only datasets were used to train the GRU model across many epochs, precisely set at 1000. However, we put an early stopping mechanism in place to avoid over-training by the models and maximize training time. With this strategy, the model stops training when the validation loss stops improving. This early stopping will help prevent overfitting and pointless calculations. Other parameters, such as the learning rate set to 0.001 with a dropout rate of 0.5, were maintained constant throughout these studies. The Adam optimizer optimized the models \cite{kingma2014adam}, trained with the Mean Squared Error Loss function, and controlled by a learning rate scheduler with a gamma of 0.1 and a step size of 50. A patience parameter of 100 epochs was used to control this further to establish the early stopping threshold. This indicates that the training process is stopped if the validation loss does not go down for 100 consecutive epochs. This is also important to guarantee that the saved model performed the best on the validation set, as shown by the lowest validation loss.

Regarding the loss function, we employed the Mean Squared Error (MSE) loss, and it is mathematically represented as equation \eqref{eq11},

\begin{equation}
    \text{MSE Loss} = \frac{1}{n} \sum_{i=1}^{n} (y_i - \hat{y}_i)^2
\label{eq11}
\end{equation}

\noindent where $y_i$ is the actual value, $\hat{y}_i$ is the predicted value, and n is the number of samples. This loss function is particularly suited for our task as it effectively captures the average squared difference between the predicted and actual trajectory points, emphasizing more significant errors.

The loss is reduced by adjusting the model's parameters through the optimizer by making forward and backward passes through the model as part of the training loop for every batch of the training data. Therefore, the Adam optimizer was employed due to its adeptness in managing sparse gradients and its capacity for adaptive learning rate. In order to improve the model's convergence, a learning rate scheduler was also used to modify the learning rate during training. We have implemented all experiments using PyTorch version 2.0.0 on a Quadro RTX 8000 GPU.

We concentrated on the model set with 128 hidden units as an extension of our training and hyperparameter tuning approach, investigating the tradeoff between performance and growing model complexity. For this analysis, the model's layer count was varied from two to ten. This investigation aimed to determine how the GRU network's capacity to learn and predict drone trajectories is impacted by its depth.

The 128 hidden unit configuration was chosen for this analysis due to its intermediate position between the most sophisticated model and the baseline. Therefore, we consider using this configuration for studying the consequences of growing complexity without going to the limits of the computing power of the largest model configuration. We observed changes in the model's learning dynamics and validation set performance by progressively increasing the number of layers. We could establish a direct correlation between model complexity and performance through hyperparameter optimization.

\subsection*{D. Analysis}

In our study, we evaluated the performance of our GRU-based trajectory predictor model using an extensive set of evaluation metrics. These measurements are essential to understanding in depth how well the model predicts drone trajectories. Each metric provides a different viewpoint on the accuracy and generalization capacity of the model.

\subsubsection{Mean Squared Error (MSE):}
The MSE is a standard metric for regression tasks, measuring the average of the squares of the errors. It is calculated as following:

\begin{equation}
    \text{MSE} = \frac{1}{n} \sum_{i=1}^{n} (y_i - \hat{y}_i)^2
\label{eq:mse}
\end{equation}

\subsubsection{Root Mean Squared Error (RMSE):}
RMSE provides a more interpretable version of MSE by taking the square root of the MSE. It is defined as:

\begin{equation}
    \text{RMSE} = \sqrt{\frac{1}{n} \sum_{i=1}^{n} (y_i - \hat{y}_i)^2}
\label{eq:rmse}
\end{equation}

\subsubsection{Mean Absolute Error (MAE):}
MAE measures the average magnitude of the errors in a set of predictions without considering their direction. It is given by:

\begin{equation}
    \text{MAE} = \frac{1}{n} \sum_{i=1}^{n} |y_i - \hat{y}_i|
\label{eq14}
\end{equation}

\subsubsection{R-squared (Coefficient of Determination):}

R-squared is a statistical measure representing the proportion of the variance for the dependent variable explained by the independent variables. It is expressed as:

\begin{equation}
    R^2 = 1 - \frac{\sum_{i=1}^{n} (y_i - \hat{y}_i)^2}{\sum_{i=1}^{n} (y_i - \overline{y})^2}
\label{eq15}
\end{equation}

\noindent where, $\overline{y}$ is the mean of the true values.

Collectively, these metrics provide a robust framework for evaluating the model's predictive performance, allowing us to assess its accuracy, error magnitude, and the proportion of variance explained by the model, which is crucial for developing effective trajectory prediction models.

\section{Results}\label{sec:results}

\subsection{Training Results}

The training phase of our study involved an in-depth evaluation of 24 GRU models. These models were distinguished by their level of complexity, the kind of trajectory data (velocity or position), the dataset (mixed or simulated), and the normalization technique (Max or Whitening) that was applied. The outcomes of this stage are essential for understanding how these factors impact the learning and generalization capacities of the model.

Table \ref{tab:ModelTrainingOutcome} overviews each model performance. The last training and validation losses are shown in this table, along with the best epoch in which early stopping was initiated. This information provides insight into the efficiency and convergence behavior of each model.

\begin{table*}[ht]
\centering
\caption{Summary of GRU Model Training Outcomes.}
\label{tab:ModelTrainingOutcome} 
\resizebox{\textwidth}{!}{%
\begin{tabular}{cccccccc}
\toprule
Model ID & Model Complexity & Trajectory Type & Dataset Type & Normalization Method & Final Training Loss & Final Validation Loss & Early Stopping Epoch \\ 
\midrule
GRU\_1  & 64 units, 2 layers  & Position & mixed & Max        & \textbf{2.2E-08} & \textbf{2.2E-08} & 380 \\
GRU\_2  & 128 units, 3 layers & Position & mixed & Max        & \textbf{2.0E-08} & \textbf{2.0E-08} & \textbf{390} \\
GRU\_3  & 256 units, 5 layers & Position & mixed & Max        & 2.4E-08 & 2.4E-08 & 346 \\ \hline
GRU\_4  & 64 units, 2 layers  & Position & mixed & Weighting  & 1.7E-05 & 1.7E-05 & 352 \\
GRU\_5  & 128 units, 3 layers & Position & mixed & Weighting  & 1.2E-05 & 1.2E-05 & 350 \\
GRU\_6  & 256 units, 5 layers & Position & mixed & Weighting  & 1.9E-04 & 1.9E-04 & 180 \\ \hline
GRU\_7  & 64 units, 2 layers  & Velocity & mixed & Max        & 9.3E-05 & 9.6E-05 & 349 \\
GRU\_8  & 128 units, 3 layers & Velocity & mixed & Max        & 4.4E-05 & 4.6E-05 & 349 \\
GRU\_9  & 256 units, 5 layers & Velocity & mixed & Max        & \textbf{1.8E-05} & \textbf{2.2E-05} & 349 \\ \hline
GRU\_10 & 64 units, 2 layers  & Velocity & mixed & Weighting  & 1.2E-02 & 1.2E-02 & 349 \\
GRU\_11 & 128 units, 3 layers & Velocity & mixed & Weighting  & 5.0E-03 & 5.6E-03 & 343 \\
GRU\_12 & 256 units, 5 layers & Velocity & mixed & Weighting  & 4.7E-02 & 2.9E-03 & 290 \\ \hline
GRU\_13 & 64 units, 2 layers  & Position & Simulation & Max        & 1.8E-06 & 1.9E-06 & 352 \\
GRU\_14 & 128 units, 3 layers & Position & Simulation & Max        & \textbf{1.4E-06} & \textbf{1.5E-06} & \textbf{390} \\
GRU\_15 & 256 units, 5 layers & Position & Simulation & Max        & \textbf{1.4E-06} & 1.6E-06 & \textbf{390} \\ \hline
GRU\_16 & 64 units, 2 layers  & Position & Simulation & Weighting  & 8.5E-05 & 9.6E-05 & 390 \\
GRU\_17 & 128 units, 3 layers & Position & Simulation & Weighting  & 4.1E-05 & 5.8E-05 & 340 \\
GRU\_18 & 256 units, 5 layers & Position & Simulation & Weighting  & 1.7E-05 & 3.6E-05 & 380 \\ \hline
GRU\_19 & 64 units, 2 layers  & Velocity & Simulation & Max        & 2.6E-04 & 2.9E-04 & 298 \\
GRU\_20 & 128 units, 3 layers & Velocity & Simulation & Max        & 1.6E-04 & 1.9E-04 & 340 \\
GRU\_21 & 256 units, 5 layers & Velocity & Simulation & Max        & \textbf{9.9E-05} & \textbf{1.4E-04} & 330 \\ \hline
GRU\_22 & 64 units, 2 layers  & Velocity & Simulation & Weighting  & 1.2E-02 & 1.4E-02 & 350 \\
GRU\_23 & 128 units, 3 layers & Velocity & Simulation & Weighting  & 5.9E-03 & 7.6E-03 & 340 \\
GRU\_24 & 256 units, 5 layers & Velocity & Simulation & Weighting  & 1.2E-02 & 1.3E-02 & 130 \\ 
\bottomrule
\end{tabular}%
}
\end{table*}

\begin{figure*}[ht]
\centering
\begin{subfigure}{.5\textwidth}
  \centering
  \includegraphics[width=.9\linewidth]{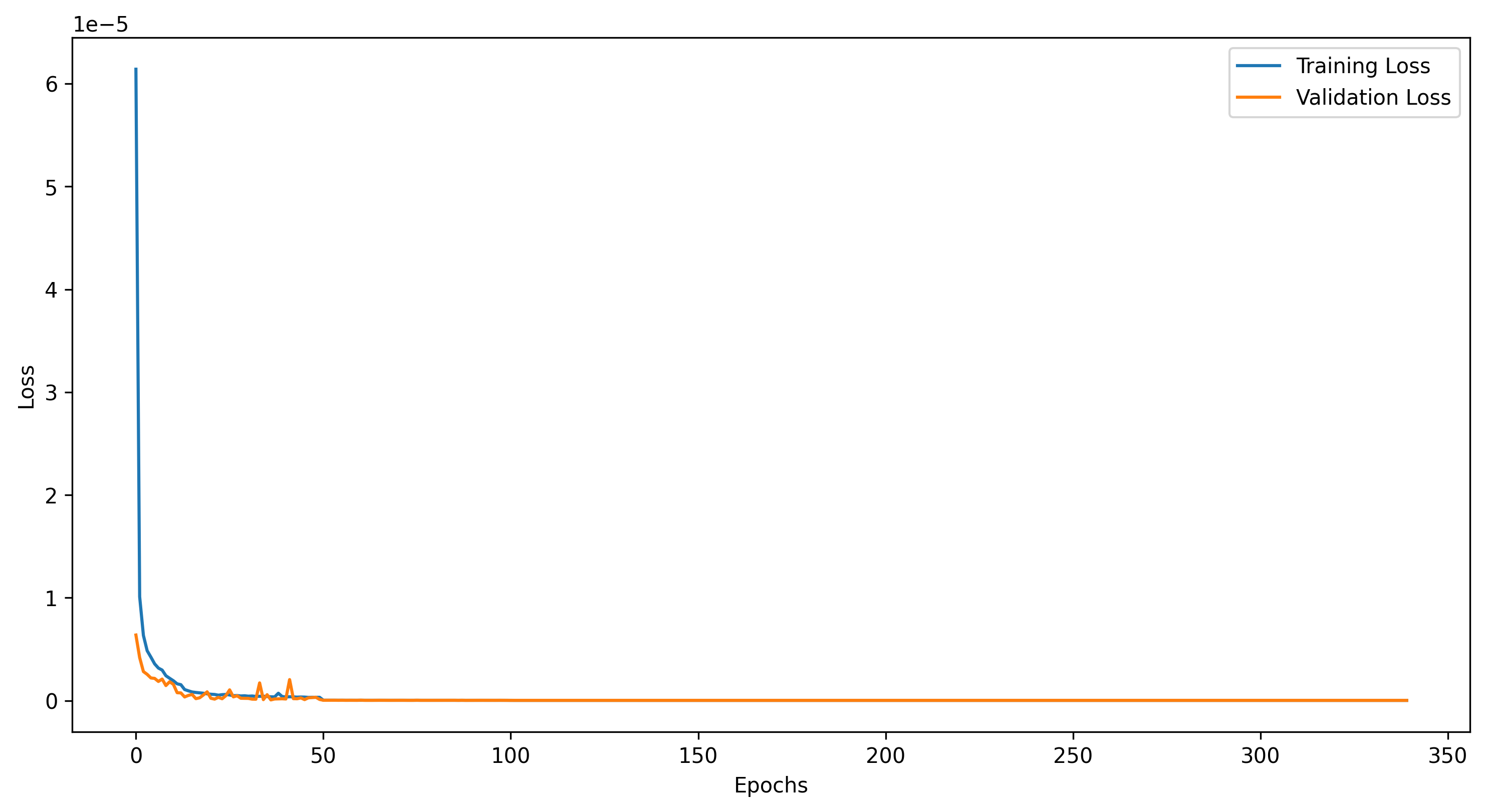}
  \caption{Training Loss with Max Norm}
  \label{fig:max_norm_train_loss}
\end{subfigure}%
\begin{subfigure}{.5\textwidth}
  \centering
  \includegraphics[width=.9\linewidth]{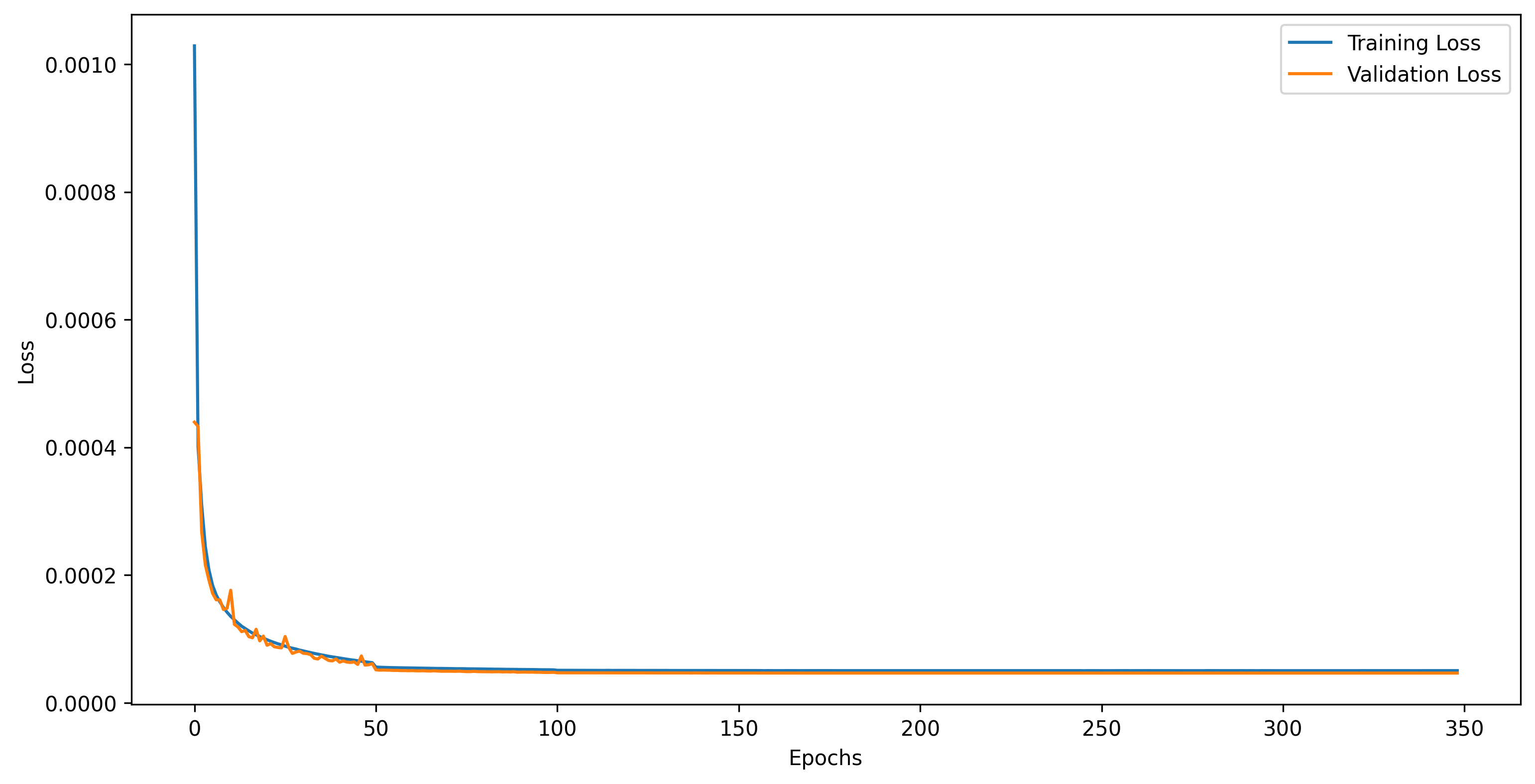}
  \caption{Validation Loss with Max Norm}
  \label{fig:max_norm_val_loss}
\end{subfigure}
\begin{subfigure}{.5\textwidth}
  \centering
  \includegraphics[width=.9\linewidth]{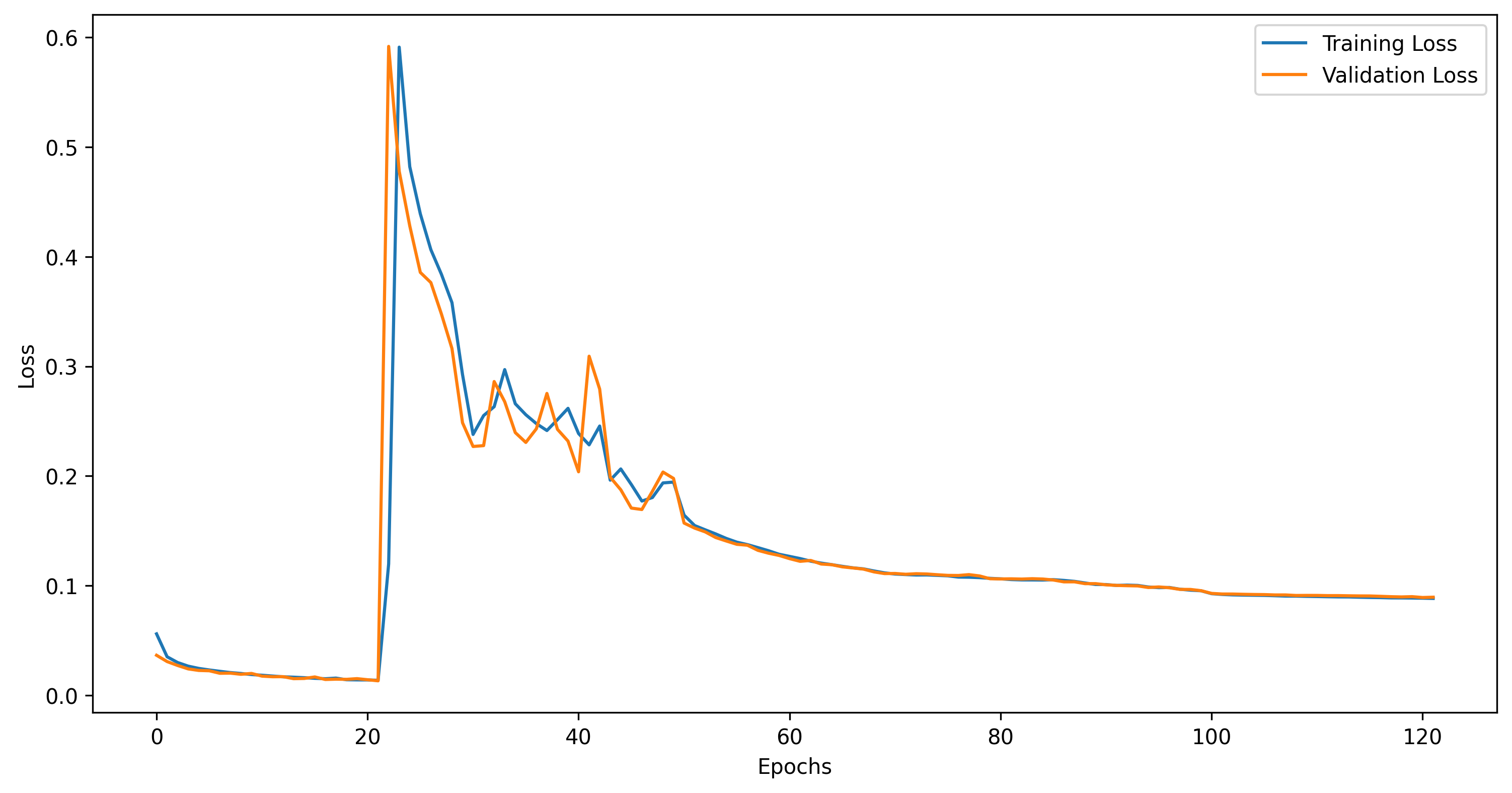}
  \caption{Training Loss with Whitening Norm}
  \label{fig:whitening_norm_train_loss}
\end{subfigure}%
\begin{subfigure}{.5\textwidth}
  \centering
  \includegraphics[width=.9\linewidth]{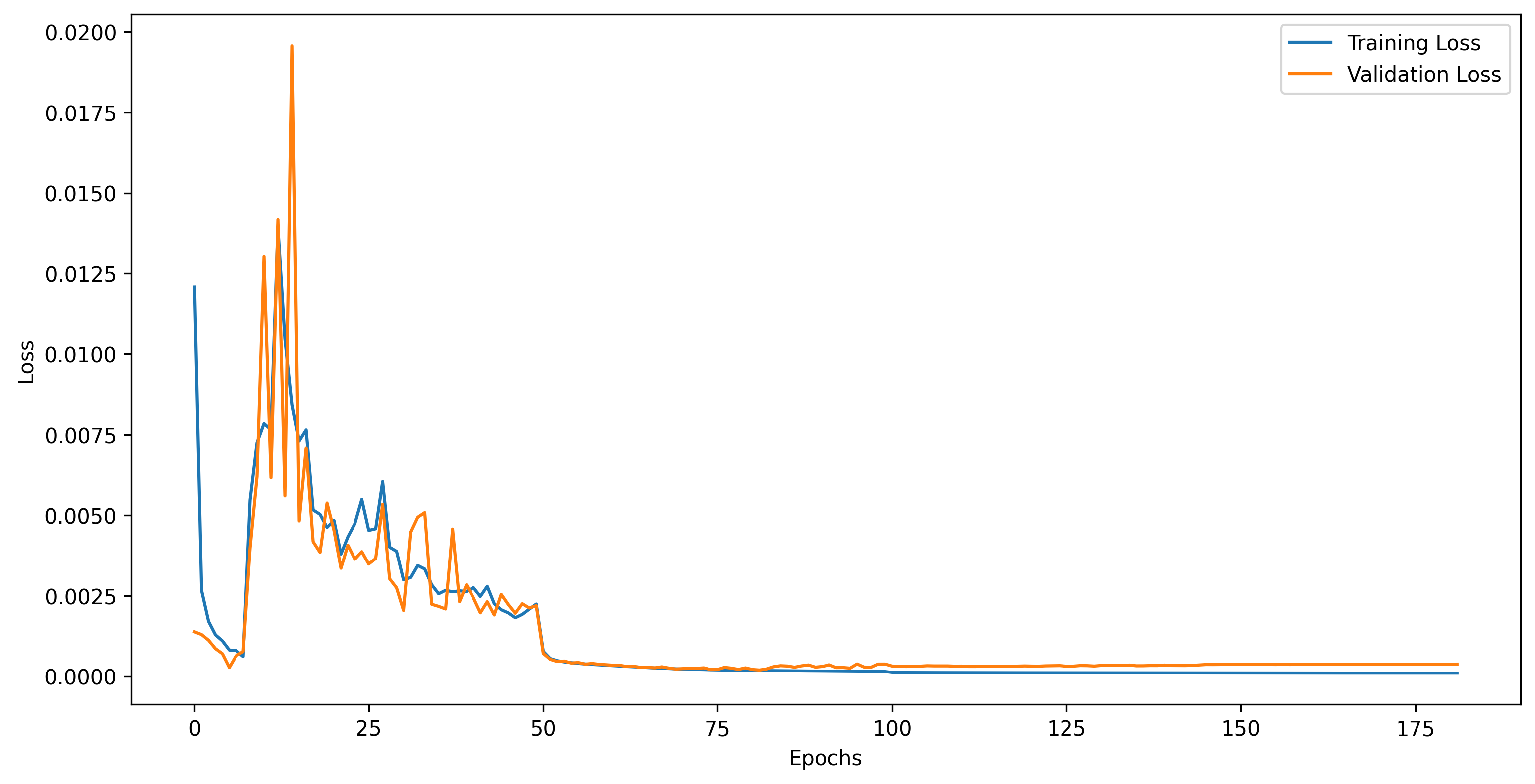}
  \caption{Validation Loss with Whitening Norm}
  \label{fig:whitening_norm_val_loss}
\end{subfigure}
\caption{Comparison of Training and Validation Losses across epochs for Max and Whitening Normalization Methods}
\label{fig:TrainingLoss}
\end{figure*}

From the summary in Table \ref{tab:ModelTrainingOutcome}, it is observed that a varying performance across models indicates the influence of model complexity, data type, and normalization methods on learning effectiveness. GRU models using the max normalization method achieved better validation loss through different models where GRU models of mixed position dataset achieved the least loss of $2 \times 10^{-8}$ across different model complexities. Additionally, the mixed velocity dataset also achieved good results for the max normalization method, indicating models normalized with the Max Norm method consistently outperformed those using the Whitening normalization across various configurations where the max normalization method might be more robust to outliers compared to Whitening normalization. Whitening, which involves linear transformations based on the covariance matrix, could be more sensitive to outliers. Furthermore, normalization affects the model's learning dynamics, an essential factor in this study. Max Norm might result in a more stable and consistent gradient flow during training, which is beneficial for the convergence of the GRU model. Whitening, on the other hand, could introduce complexities in the data that make it harder for the model to learn effectively. Figure \ref{fig:TrainingLoss} shows the training and validation loss for the training epochs for both Max and Whitening normalization techniques. 

As shown in Fig \ref{fig:TrainingLoss}, the GRU models normalized using the Max Norm and Whitening approaches exhibit distinct differences in convergence. Figure \ref{fig:TrainingLoss} (a) and (b) demonstrate the Max Norm models, which are steady and smooth convergence, suggesting a reliable learning mechanism and efficient generalization. On the other hand, loss graphs of models using Whitening normalization shown in (c) and (d) show notable oscillations, indicating a less consistent and more unpredictable learning process. This distinction emphasizes the importance of selecting the appropriate normalization method for better model training.

Other insights are gained from the summary table \ref{tab:ModelTrainingOutcome} focusing on training different datasets, including mixed datasets (short, long, simulation) and drone trajectories versus only simulation datasets. Table \ref{tab:ModelTrainingOutcome} indicates a general trend where models trained on the mixed dataset consistently outperform those trained on simulation data. This observation suggests that the mixed dataset, which combines short, long, and simulated data, provides a more diverse and comprehensive learning environment. Due to this diversity, the models are probably exposed to a greater variety of scenarios and variations in drone trajectories, which improves their capacity to generalize on data that has not been observed.

\subsection{Test Set Results}

\begin{table*}[ht]
\centering
\caption{Performance Metrics for Drone Trajectory Simulations under Various Model Complexities.}
\label{tab:Performance Metrics}
\resizebox{\textwidth}{!}{%
\begin{tabular}{|ccccccccc|}
\hline
\multicolumn{9}{|c|}{Hidden dimensions 64 with Number of layer 2} \\ \hline
\multicolumn{5}{|c|}{Simulation Drone Traj} &
  \multicolumn{4}{c|}{Mix Drone Traj} \\ \hline
\multicolumn{1}{|c|}{} &
  \multicolumn{2}{c|}{Position Points} &
  \multicolumn{2}{c|}{Velocity Points} &
  \multicolumn{2}{c|}{Position Points} &
  \multicolumn{2}{c|}{Velocity Points} \\ \hline
\multicolumn{1}{|c|}{Metrics} &
  \multicolumn{1}{c|}{Max Norm} &
  \multicolumn{1}{c|}{Whiting Norm} &
  \multicolumn{1}{c|}{Max Norm} &
  \multicolumn{1}{c|}{Whiting Norm} &
  \multicolumn{1}{c|}{Max Norm} &
  \multicolumn{1}{c|}{Whiting Norm} &
  \multicolumn{1}{c|}{Max Norm} &
  Whiting Norm \\ \hline
\multicolumn{1}{|c|}{MSE} &
  \multicolumn{1}{c|}{1.91E-06} &
  \multicolumn{1}{c|}{9.66E-05} &
  \multicolumn{1}{c|}{2.87E-04} &
  \multicolumn{1}{c|}{1.39E-02} &
  \multicolumn{1}{c|}{\textbf{2.20E-08}} &
  \multicolumn{1}{c|}{1.71E-05} &
  \multicolumn{1}{c|}{9.63E-05} &
  1.23E-02 \\ \hline
\multicolumn{1}{|c|}{RMSE} &
  \multicolumn{1}{c|}{1.37E-03} &
  \multicolumn{1}{c|}{9.71E-03} &
  \multicolumn{1}{c|}{1.69E-02} &
  \multicolumn{1}{c|}{1.17E-01} &
  \multicolumn{1}{c|}{\textbf{1.46E-04}} &
  \multicolumn{1}{c|}{4.09E-03} &
  \multicolumn{1}{c|}{9.74E-03} &
  1.10E-01 \\ \hline
\multicolumn{1}{|c|}{MAE} &
  \multicolumn{1}{c|}{8.30E-04} &
  \multicolumn{1}{c|}{4.94E-03} &
  \multicolumn{1}{c|}{1.17E-02} &
  \multicolumn{1}{c|}{7.96E-02} &
  \multicolumn{1}{c|}{\textbf{8.42E-05}} &
  \multicolumn{1}{c|}{1.99E-03} &
  \multicolumn{1}{c|}{6.20E-03} &
  6.39E-02 \\ \hline
\multicolumn{1}{|c|}{R2} &
  \multicolumn{1}{c|}{\textbf{1.00E+00}} &
  \multicolumn{1}{c|}{\textbf{1.00E+00}} &
  \multicolumn{1}{c|}{9.92E-01} &
  \multicolumn{1}{c|}{9.86E-01} &
  \multicolumn{1}{c|}{\textbf{1.00E+00}} &
  \multicolumn{1}{c|}{\textbf{1.00E+00}} &
  \multicolumn{1}{c|}{9.95E-01} &
  9.88E-01 \\ \hline
\multicolumn{1}{|c|}{} &
  \multicolumn{8}{c|}{Hidden dimensions 128 with Number of layer 3} \\ \hline
\multicolumn{1}{|c|}{Metrics} &
  \multicolumn{1}{c|}{Max Norm} &
  \multicolumn{1}{c|}{Whiting Norm} &
  \multicolumn{1}{c|}{Max Norm} &
  \multicolumn{1}{c|}{Whiting Norm} &
  \multicolumn{1}{c|}{Max Norm} &
  \multicolumn{1}{c|}{Whiting Norm} &
  \multicolumn{1}{c|}{Max Norm} &
  Whiting Norm \\ \hline
\multicolumn{1}{|c|}{MSE} &
  \multicolumn{1}{c|}{1.52E-06} &
  \multicolumn{1}{c|}{5.91E-05} &
  \multicolumn{1}{c|}{1.91E-04} &
  \multicolumn{1}{c|}{7.67E-03} &
  \multicolumn{1}{c|}{\textbf{2.10E-08}} &
  \multicolumn{1}{c|}{1.21E-05} &
  \multicolumn{1}{c|}{4.70E-05} &
  5.69E-03 \\ \hline
\multicolumn{1}{|c|}{RMSE} &
  \multicolumn{1}{c|}{1.23E-03} &
  \multicolumn{1}{c|}{7.59E-03} &
  \multicolumn{1}{c|}{1.38E-02} &
  \multicolumn{1}{c|}{8.73E-02} &
  \multicolumn{1}{c|}{\textbf{1.42E-04}} &
  \multicolumn{1}{c|}{3.44E-03} &
  \multicolumn{1}{c|}{6.83E-03} &
  7.49E-02 \\ \hline
\multicolumn{1}{|c|}{MAE} &
  \multicolumn{1}{c|}{7.49E-04} &
  \multicolumn{1}{c|}{4.12E-03} &
  \multicolumn{1}{c|}{9.98E-03} &
  \multicolumn{1}{c|}{6.15E-02} &
  \multicolumn{1}{c|}{\textbf{8.13E-05}} &
  \multicolumn{1}{c|}{1.70E-03} &
  \multicolumn{1}{c|}{4.66E-03} &
  4.74E-02 \\ \hline
\multicolumn{1}{|c|}{R2} &
  \multicolumn{1}{c|}{\textbf{1.00E+00}} &
  \multicolumn{1}{c|}{\textbf{1.00E+00}} &
  \multicolumn{1}{c|}{9.95E-01} &
  \multicolumn{1}{c|}{9.92E-01} &
  \multicolumn{1}{c|}{\textbf{1.00E+00}} &
  \multicolumn{1}{c|}{\textbf{1.00E+00}} &
  \multicolumn{1}{c|}{9.97E-01} &
  9.95E-01 \\ \hline
\multicolumn{1}{|c|}{} &
  \multicolumn{8}{c|}{Hidden dimensions 256 with Number of layer 5} \\ \hline
\multicolumn{1}{|c|}{Metrics} &
  \multicolumn{1}{c|}{Max Norm} &
  \multicolumn{1}{c|}{Whiting Norm} &
  \multicolumn{1}{c|}{Max Norm} &
  \multicolumn{1}{c|}{Whiting Norm} &
  \multicolumn{1}{c|}{Max Norm} &
  \multicolumn{1}{c|}{Whiting Norm} &
  \multicolumn{1}{c|}{Max Norm} &
  Whiting Norm \\ \hline
\multicolumn{1}{|c|}{MSE} &
  \multicolumn{1}{c|}{1.57E-06} &
  \multicolumn{1}{c|}{3.71E-05} &
  \multicolumn{1}{c|}{1.40E-04} &
  \multicolumn{1}{c|}{1.35E-02} &
  \multicolumn{1}{c|}{\textbf{2.50E-08}} &
  \multicolumn{1}{c|}{1.95E-04} &
  \multicolumn{1}{c|}{2.17E-05} &
  2.95E-03 \\ \hline
\multicolumn{1}{|c|}{RMSE} &
  \multicolumn{1}{c|}{1.25E-03} &
  \multicolumn{1}{c|}{5.98E-03} &
  \multicolumn{1}{c|}{1.18E-02} &
  \multicolumn{1}{c|}{1.15E-01} &
  \multicolumn{1}{c|}{\textbf{1.55E-04}} &
  \multicolumn{1}{c|}{1.39E-02} &
  \multicolumn{1}{c|}{4.65E-03} &
  5.39E-02 \\ \hline
\multicolumn{1}{|c|}{MAE} &
  \multicolumn{1}{c|}{7.60E-04} &
  \multicolumn{1}{c|}{3.45E-03} &
  \multicolumn{1}{c|}{8.73E-03} &
  \multicolumn{1}{c|}{7.72E-02} &
  \multicolumn{1}{c|}{\textbf{9.03E-05}} &
  \multicolumn{1}{c|}{9.49E-03} &
  \multicolumn{1}{c|}{3.26E-03} &
  3.55E-02 \\ \hline
\multicolumn{1}{|c|}{R2} &
  \multicolumn{1}{c|}{\textbf{1.00E+00}} &
  \multicolumn{1}{c|}{\textbf{1.00E+00}} &
  \multicolumn{1}{c|}{9.96E-01} &
  \multicolumn{1}{c|}{9.86E-01} &
  \multicolumn{1}{c|}{\textbf{1.00E+00}} &
  \multicolumn{1}{c|}{\textbf{1.00E+00}} &
  \multicolumn{1}{c|}{9.99E-01} &
  9.97E-01 \\ \hline
\end{tabular}%
}
\end{table*}

In this work, we investigate and evaluate the effect of model complexity on different drone trajectory test datasets, including both simulated and mixed datasets, which feature short and long trajectories with various normalization methods. As shown in Table \ref{tab:Performance Metrics}, performance improves consistently across all evaluation metrics as model complexity increases—from 64 hidden dimensions with two layers to 256 dimensions with five layers. This improvement suggests that more complex models are better at capturing the intricate movement patterns of drone trajectories. Among the evaluated metrics, including MSE, RMSE, MAE, and R², increasing model complexity leads to a significant reduction in MSE and RMSE, which indicates improved prediction accuracy (i.e., the predicted values are closer to the true values). Additionally, the decline in MAE further emphasizes greater prediction precision, understood here as the consistency of predictions across different test points. Across all experiments, the models show consistent performance for both position and velocity data. However, error rates in velocity predictions are generally higher compared to those in position predictions, reflecting the greater difficulty in accurately predicting velocity changes in dynamic UAV movements.

\begin{figure*}[ht]
\centering

\begin{subfigure}{.30\textwidth}
  \centering
  \includegraphics[width=.95\linewidth]{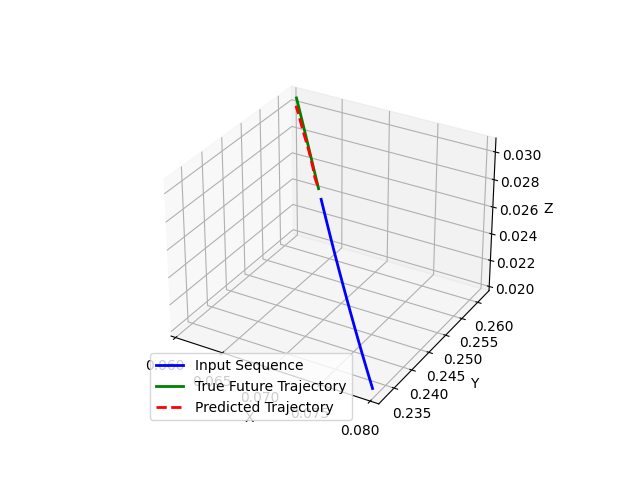}
  \caption{Position with Max (GRU\_3)}
  \label{fig:model_3_combined}
\end{subfigure}%
\begin{subfigure}{.30\textwidth}
  \centering
  \includegraphics[width=.95\linewidth]{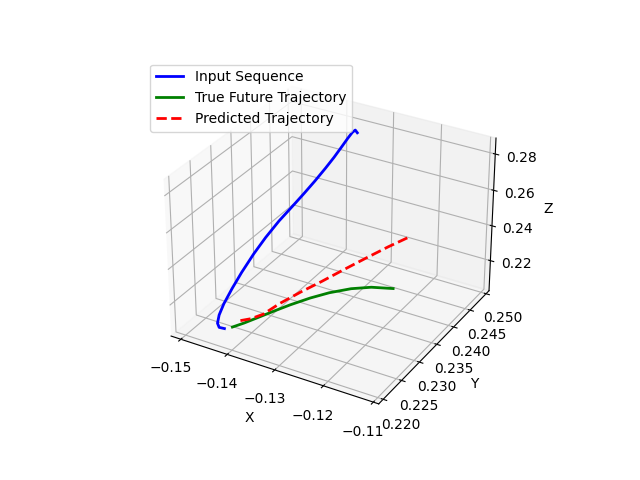}
  \caption{Position with Whitening (GRU\_5)}
  \label{fig:model_5_combined}
\end{subfigure}%
\begin{subfigure}{.30\textwidth}
  \centering
  \includegraphics[width=.95\linewidth]{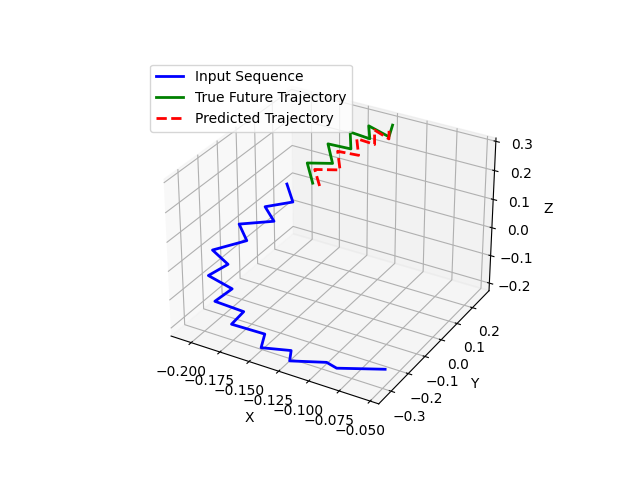}
  \caption{Velocity with Max (GRU\_10)}
  \label{fig:model_17_combined}
\end{subfigure}

\begin{subfigure}{.30\textwidth}
  \centering
  \includegraphics[width=.95\linewidth]{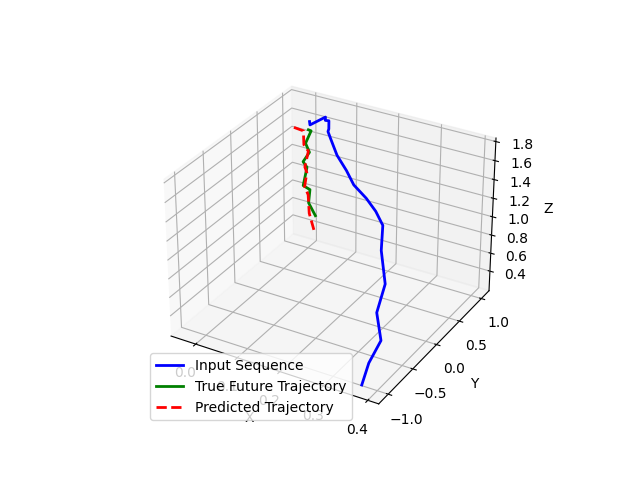}
  \caption{Velocity with Whitening (GRU\_12)}
  \label{fig:model_15_combined}
\end{subfigure}%
\begin{subfigure}{.30\textwidth}
  \centering
  \includegraphics[width=.95\linewidth]{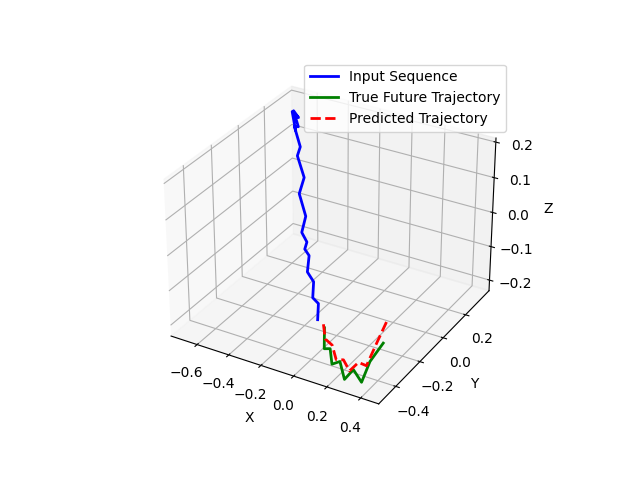}
  \caption{Velocity with Whitening (GRU\_22)}
  \label{fig:model_16_combined}
\end{subfigure}%
\begin{subfigure}{.30\textwidth}
  \centering
  \includegraphics[width=.95\linewidth]{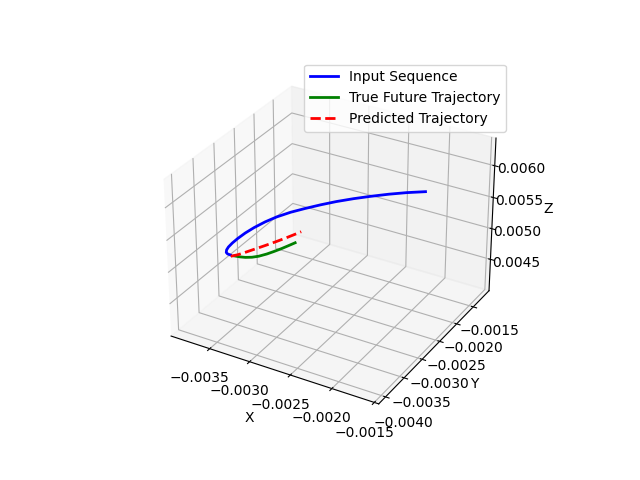}
  \caption{Position with Max (GRU\_1)}
  \label{fig:model_1_combined}
\end{subfigure}

\caption{Comparison of Input, Target, and Predicted Trajectories across Position and Velocity Datasets with Max and Whitening Normalization. Includes models trained on simulation-only and mixed datasets.}
\label{fig:CombinedTrajPred}
\end{figure*}

On the other hand, different normalization methods, including the Max and Whitening normalization techniques for all models, have been trained and tested to investigate GRU models' learning and prediction behavior. As shown in Table \ref{tab:Performance Metrics}, performance indicators consistently improve for the Max normalization technique. In particular, using the Max Normalization method, The Position Mix dataset achieved an average MSE score of $2\times 10^{-8}$ among all model complexities. Furthermore, the results of the drone trajectories based on velocity prediction improved using the Max normalization by an average of 70\%  compared with the Whitening normalization method. Generally, the Max Norm typically provides more notable enhancements. Figure \ref{fig:CombinedTrajPred} shows a comparison of input, target, and predicted trajectories across position and velocity datasets using Max and Whitening Normalization techniques. The figure includes results from GRU models of varying complexity, trained on both simulation-only and mixed datasets (real and simulated data). In the first row, (a) Position with Max (GRU\_3), with 256 hidden dimensions, demonstrates a close alignment between predicted and target trajectories, indicating high precision. In contrast, (b) Position with Whitening (GRU\_5), with 128 hidden dimensions, shows noticeable deviations, suggesting that Whitening Normalization is less effective for complex trajectory predictions and (c) Velocity with Max (GRU\_10), with 128 hidden dimensions, shows strong performance in predicting dynamic changes in velocity, aligning closely with the target.

In the second row, (d) Velocity with Whitening (GRU\_12), with 256 hidden dimensions, shows significant deviations, similar to the position data. (e) Velocity with Whitening (GRU\_22), trained on simulation-only data, captures basic movement patterns but lacks the accuracy seen in (f) Position with Max (GRU\_1), which uses Max Normalization and mixed data, leading to more accurate predictions, even with 64 hidden dimensions.

As shown in Figure~\ref{fig:CombinedTrajPred}, Max Normalization consistently outperforms Whitening Normalization for both position and velocity data, leading to more accurate trajectory predictions, particularly in dynamic and complex flight paths. Velocity data also proves more effective in capturing the essence of UAV movement dynamics, as velocity better reflects the variations in speed and direction. Furthermore, models trained on mixed datasets provide more robust predictions compared to those trained on simulation-only data, underscoring the importance of including real-world data in training to improve generalization and accuracy. Model complexity (number of hidden dimensions) also plays a role, with higher complexity models generally producing more precise predictions, though simpler models with Max Normalization still perform adequately when trained on mixed data.

Our experimental findings indicate that velocity data, as compared to position data segments, has shown to be more effective in drone trajectory prediction. The reliable, predictable range of velocity data captures the essence of the drone's movement dynamics, including variations in both direction and speed. This consistency in data distribution is beneficial when training models, especially for architectures that are good at handling temporal sequences, such as Gated Recurrent Units (GRUs). Additionally, to make the training process convergence better, the velocity data has a constant range, which helps make the predictive models more broadly applicable to a broader range of drone flight circumstances. By ensuring that the models remain unaffected by the spatial extremities that may be present in positioning data, this uniformity helps to keep the focus on learning the fundamental movement patterns. Our study's empirical results support this understanding, which shows that models trained on velocity data consistently beat those trained on position data across a wide range of essential performance criteria. This finding confirms that velocity data is crucial in accurately predicting UAV trajectories and highlights its applicability in applications requiring temporal accuracy and dynamic understanding.

Moreover, instead of letting the prediction models tend towards complex architectures, we penalize the increased complexity using the adjusted R2 instead of the standard one. In this way, we can compare all models on equal feet. Therefore, varying the number of hidden layers from 2 to 10 is not biased in the reported results. The modified or adjusted R2 is defined below:

\begin{eqnarray}
\text{Adjusted-R}^2 = \frac{R^2}{128 + n} 
\end{eqnarray}

where $ n $ represents the number of layers used.

\begin{figure}[ht]
    \centering
    \includegraphics[width=0.7\textwidth]{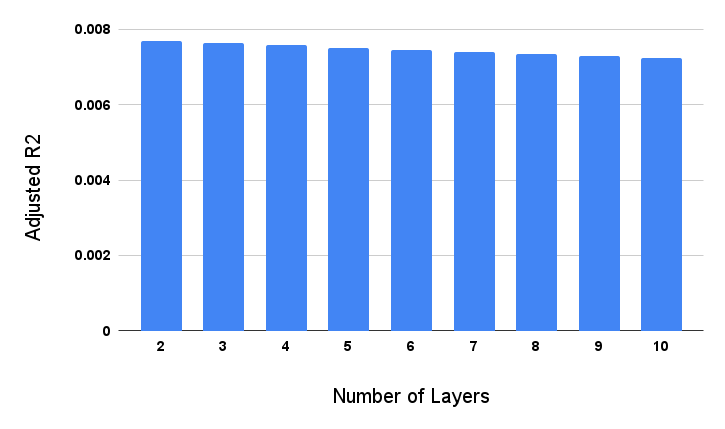}
    \caption{Adjusted R-squared for GRU models with 2 to 10 layers, showing the effect of model complexity on performance.}
    \label{fig:adjustedr2}
\end{figure}

Figures \ref{fig:adjustedr2} support our claims as there is a slight decrease in the R2 scores to penalize the increased complexity and restrict the models from overfitting.

Furthermore, we evaluate our GRU models' performance against three other well-known recurrent neural network architectures: Bidirectional Long Short-Term Memory (BiLSTM) \cite{zhang2015bidirectional}, Long Short-Term Memory (LSTM) \cite{hochreiter1997long} and Bidirectional GRU model \cite{yang2020real}. These models serve as valuable reference points for our investigation and are benchmarks for sequential data processing in the literature. This study compares models trained on three distinct datasets: mixed position data, mixed velocity data, and simulation-only position data. All models were trained using the Max normalization approach, which is considered the best approach in our study, as illustrated in table \ref{tab:Performance Metrics}.

\begin{table*}[ht]
\centering
\caption{Comparison of LSTM, BiLSTM, BiGRU, and GRU Model Performance Across Various Datasets}
\label{tab:BenchMakring}
\resizebox{\textwidth}{!}{%
\begin{tabular}{|c|cccc|cccc|cccc|}
\hline
Data Type &
  \multicolumn{4}{c|}{mixed Position Dataset} &
  \multicolumn{4}{c|}{mixed Velocity Dataset} &
  \multicolumn{4}{c|}{Simulation Position Dataset} \\ \hline
Metrics\textbackslash Models &
  \multicolumn{1}{c|}{LSTM} &
  \multicolumn{1}{c|}{BiLSTM} &
  \multicolumn{1}{c|}{BiGRU} &
  GRU &
  \multicolumn{1}{c|}{LSTM} &
  \multicolumn{1}{c|}{BiLSTM} &
  \multicolumn{1}{c|}{BiGRU} &
  GRU &
  \multicolumn{1}{c|}{LSTM} &
  \multicolumn{1}{c|}{BiLSTM} &
  \multicolumn{1}{c|}{BiGRU} &
  GRU \\ \hline
MSE &
  \multicolumn{1}{c|}{4.5E-08} &
  \multicolumn{1}{c|}{2.8E-07} &
  \multicolumn{1}{c|}{4.9E-03} &
  \textbf{2.2E-08} &
  \multicolumn{1}{c|}{1.0E-04} &
  \multicolumn{1}{c|}{3.7E-05} &
  \multicolumn{1}{c|}{1.9E-02} &
  \textbf{2.5E-08} &
  \multicolumn{1}{c|}{5.1E-06} &
  \multicolumn{1}{c|}{4.5E-06} &
  \multicolumn{1}{c|}{9.3E-02} &
  \textbf{1.9E-06} \\ \hline
RMSE &
  \multicolumn{1}{c|}{2.1E-04} &
  \multicolumn{1}{c|}{5.3E-04} &
  \multicolumn{1}{c|}{3.9E-02} &
  \textbf{1.5E-04} &
  \multicolumn{1}{c|}{1.0E-02} &
  \multicolumn{1}{c|}{6.1E-03} &
  \multicolumn{1}{c|}{1.4E-01} &
  \textbf{1.5E-04} &
  \multicolumn{1}{c|}{2.2E-03} &
  \multicolumn{1}{c|}{2.1E-03} &
  \multicolumn{1}{c|}{3.1E-01} &
  \textbf{1.4E-03} \\ \hline
MAE &
  \multicolumn{1}{c|}{1.4E-04} &
  \multicolumn{1}{c|}{3.3E-04} &
  \multicolumn{1}{c|}{3.0E-02} &
  \textbf{8.4E-05} &
  \multicolumn{1}{c|}{6.2E-03} &
  \multicolumn{1}{c|}{4.2E-03} &
  \multicolumn{1}{c|}{8.0E-02} &
  \textbf{9.0E-05} &
  \multicolumn{1}{c|}{1.4E-03} &
  \multicolumn{1}{c|}{1.4E-03} &
  \multicolumn{1}{c|}{2.3E-01} &
  \textbf{8.3E-04} \\ \hline
\end{tabular}%
}
\end{table*}

As shown in Table \ref{tab:BenchMakring}, with the lowest MSE, GRU models perform better than other models, especially in the mixed position and velocity datasets, demonstrating a high level of accuracy and precision. This indicates that GRUs are efficient for UAV trajectory prediction due to their less complex structure when compared to LSTMs, BiLSTMs, and BiGRU. Despite their excellent performance, the LSTM models fall short of the GRU models' accuracy. Similarly, the BiLSTM models perform comparably, particularly when capturing the data dynamics, but they are still not as good as the GRU.

However, it is essential to note that BiGRU models do not consistently outperform the simpler GRU models despite their more sophisticated architecture. Due to their bidirectional nature, BiGRUs have the potential to understand the data context better, but we found out that their performance varies across different datasets. In some cases, the additional complexity of BiGRUs does not reflect a proportional increase in prediction accuracy, indicating a trade-off between model complexity and performance efficiency.

Overall, our findings show that GRU models play a significant role in precisely forecasting drone trajectories and that different model complexities may well handle extreme values and variability in simulation data.

\subsection{Real-time Simulation Results}
\label{sec:sim_results}

For assessing our GRU model's performance in live scenarios, we established a simulation environment using Gazebo, an open-source robotics simulator \cite{koenig2004design}, in combination with PX4 autopilot and the Robot Operating System (ROS2) to streamline the software integration. The roles of these elements are detailed below.

\begin{figure}
\centering
\includegraphics[width=0.45\textwidth]{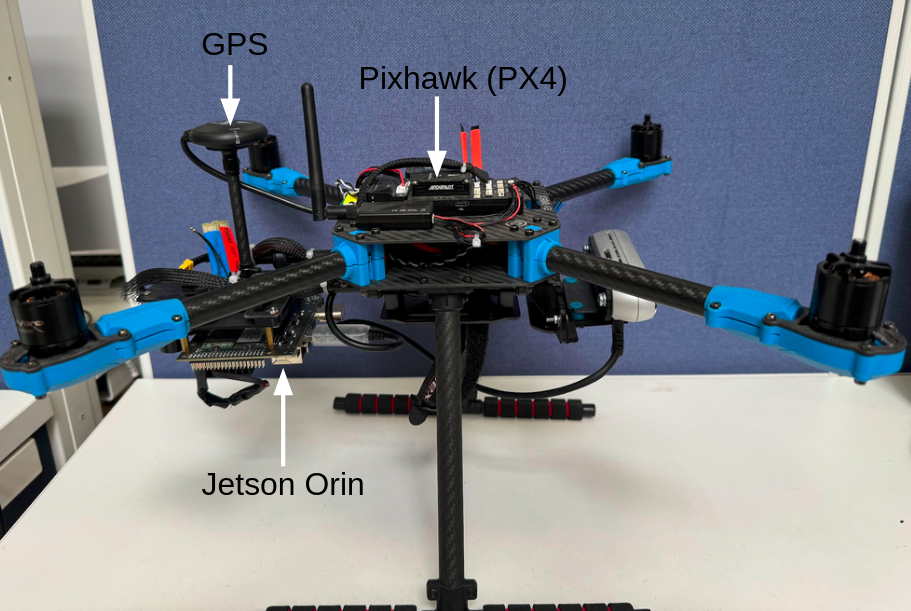}
\caption{Holybro X500 V2 Quadcopter.}
\label{fig:HolybroQuadcopter}
\end{figure}

\begin{itemize}
\item \textbf{Gazebo}: This widely recognized open-source robotics simulator tha can replicate the physical properties, sensors, and actuators of multi-rotor UAVs. Gazebo's extensive capabilities include simulating various world and robot models and environmental factors like wind that impact the robot's dynamics. We employed a Gazebo model representing a commercially available quadcopter, specifically the Holybro X500, as depicted in Figure \ref{fig:HolybroQuadcopter}, which we also use in our real-world experiments in the next section.

\item \textbf{PX4 autopilot}: The PX4 autopilot firmware \cite{7140074}, a well-established open-source system, is used for autonomous control of diverse mobile robotics, including multi-rotor UAVs and unmanned ground and surface vehicles. Its software-in-the-loop (SITL) integration with Gazebo allows for the testing of autopilot functionalities in a simulated environment before actual deployment. Gazebo simulates essential navigational sensors for the quadcopter, such as the IMU, barometer, magnetometer, and GPS. These are passed to PX4 to be processed for state estimation using an Extended Kalman Filter. Then, PX4 generates control commands for motor speed based on the desired trajectory and sends them back to Gazebo for actuation.
\item \textbf{ROS2}: The Robot Operating System (ROS 2) \cite{macenski2022robot} is employed to integrate the components mentioned above seamlessly. We interface the PX4 firmware with ROS2 using the MAVROS package \footnote{\url{https://github.com/mavlink/mavros}}, enabling the efficient exchange of data between the simulated quadcopter in Gazebo and the PX4 autopilot. The advantage of using ROS2 lies in its compatibility, allowing the transfer of the same software framework to actual systems with minimal modifications.
\end{itemize}

In our simulations, the quadcopter is instructed to follow a lemniscate (infinity-like) position trajectory, exhibiting segments of varying curvature. The trajectory's characteristics, Table \ref{tab:ParametersInfinity}, include its central point, radius, and spatial orientation as explained in section \ref{sec:dataset}.

\begin{table}[ht]
\centering
\caption{Parameters of the lemniscate trajectory in simulation experiments.}
\label{tab:traj_params}
\begin{tabular}{|l|c|}
\hline
\textbf{Parameter}    & \textbf{Value} \\
\hline
Center, $\mathbf{c}$ (m)               & $[-100.0, 0.0, 10.0]^T$ \\
\hline
Normal Vector, $\mathbf{n}$ (m)         & $[1.0, 1.0, 1.0]^T$ \\
\hline
Radius, $r$ (m)               & $3.0$ m \\
\hline
Speed, $\omega$ (rad/s)                 & $0.8$ rad/s \\
\hline
\end{tabular}
\label{tab:ParametersInfinity}
\end{table}

\begin{table*}[ht]
\centering
\caption{Average RMSE of position and velocity-based trajectory prediction in simulation experiments}
\label{tab:sim_rmse}
\resizebox{\textwidth}{!}{
\begin{tabular}{|c|c|}
\hline
Model & Average RMSE \\ \hline
Position prediction with max norm in Figure \ref{fig:sim_pos_max_norm} & 46.241 \\ \hline
Position prediction with whitening normalization Figure \ref{fig:sim_pos_white_norm}& 19.73 \\ \hline
Position prediction based on velocity model with whitening normalization Figure \ref{fig:sim_vel_white_norm}& 0.424 \\ \hline
Position prediction based on velocity model with max norm Figure \ref{fig:sim_vel_max_norm} & 0.349 \\ \hline
\end{tabular}
}
\end{table*}

To create input trajectory segments for the GRU models during simulation runs, we initially buffer a sufficient number of position samples, provided by the PX4 autopilot's state estimation system, corresponding to a 2-second duration, which is the length expected by the trained models. Given the real-time nature of the simulation, these position samples may have varied sampling times. It is, therefore, essential to resample them at the frequency used for training the GRU models, specifically $0.1s$, to generate 20 samples for the GRU model input sequence. Following this, we perform inference to yield predicted trajectory segments – 10 samples equating to a 1-second prediction. We then accumulate 100 predicted trajectory segments alongside their corresponding actual segments to compute the average errors, see \eqref{eq:rmse}.

\begin{figure*}[ht]
\centering
\begin{subfigure}{.45\textwidth}
  \centering
  \includegraphics[width=.95\linewidth]{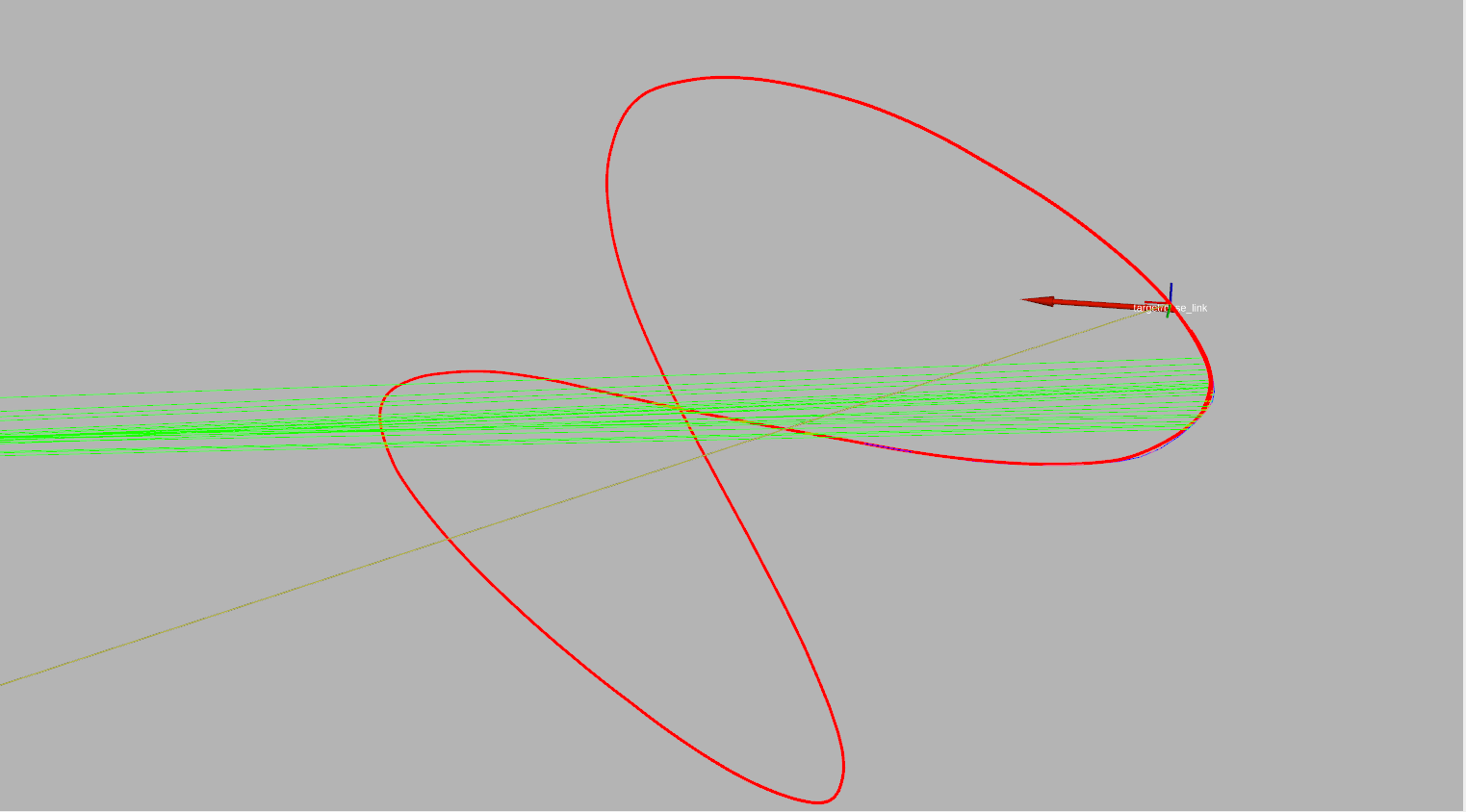}
  \caption{Trajectory prediction based on position model with max norm.}
  \label{fig:sim_pos_max_norm}
\end{subfigure}%
\begin{subfigure}{.45\textwidth}
  \centering
  \includegraphics[width=.95\linewidth]{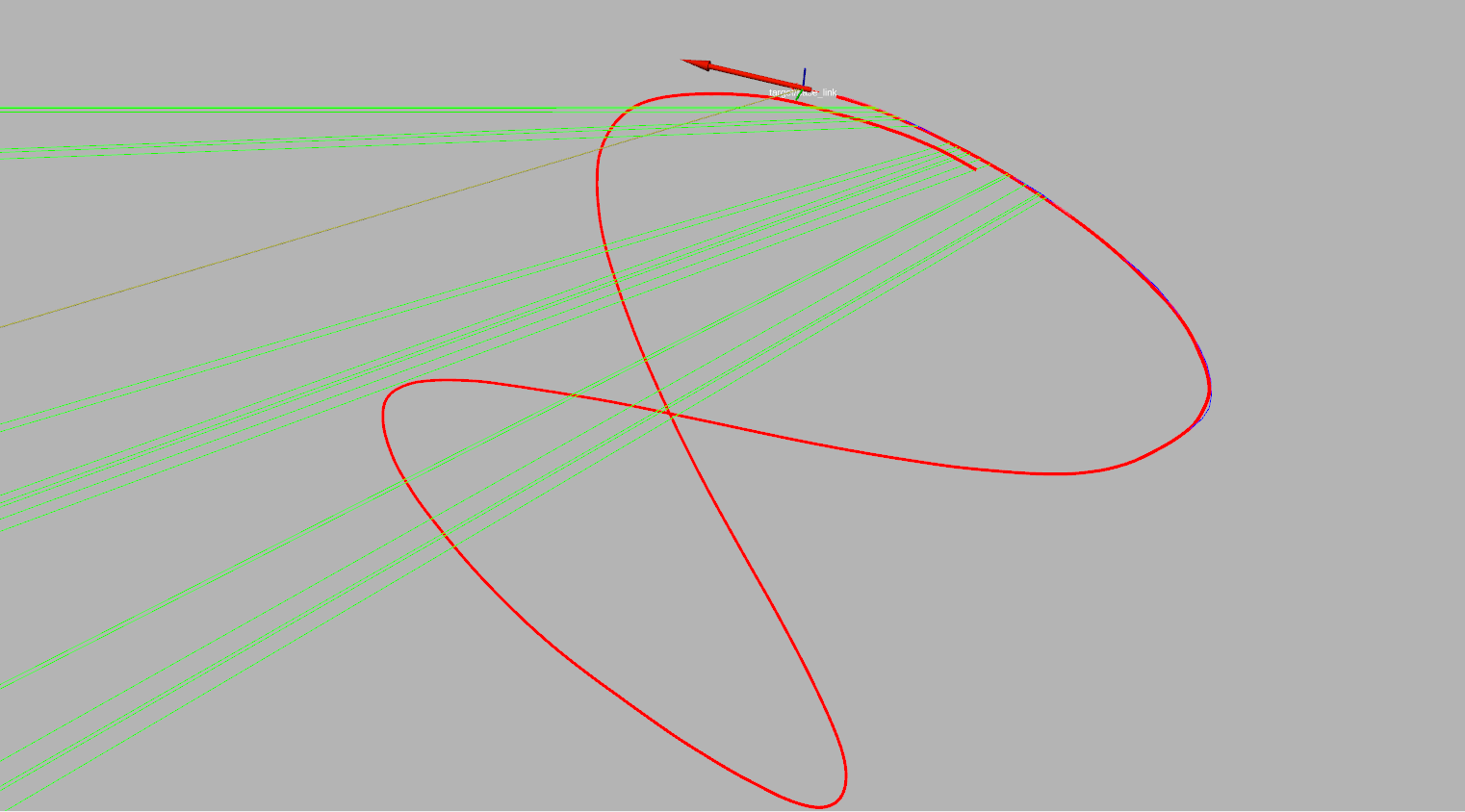}
  \caption{Trajectory prediction based on position model with whitening normalization.}
  \label{fig:sim_pos_white_norm}
\end{subfigure}%
\\ 

\begin{subfigure}{.45\textwidth}
  \centering
  \includegraphics[width=.95\linewidth]{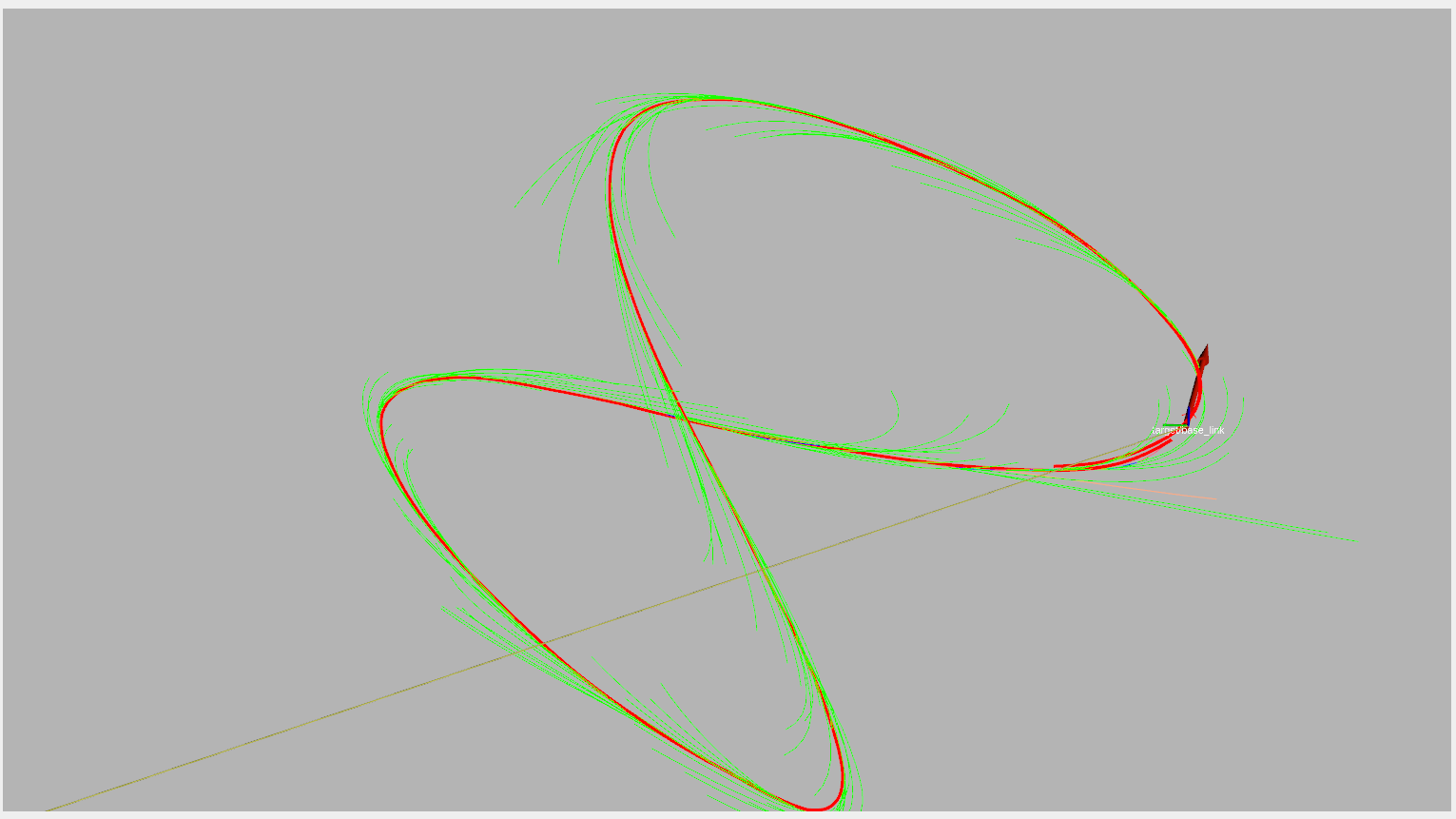}
  \caption{Trajectory prediction based on velocity model with whitening normalization.}
  \label{fig:sim_vel_white_norm}
\end{subfigure}%
\begin{subfigure}{.45\textwidth}
  \centering
  \includegraphics[width=.95\linewidth]{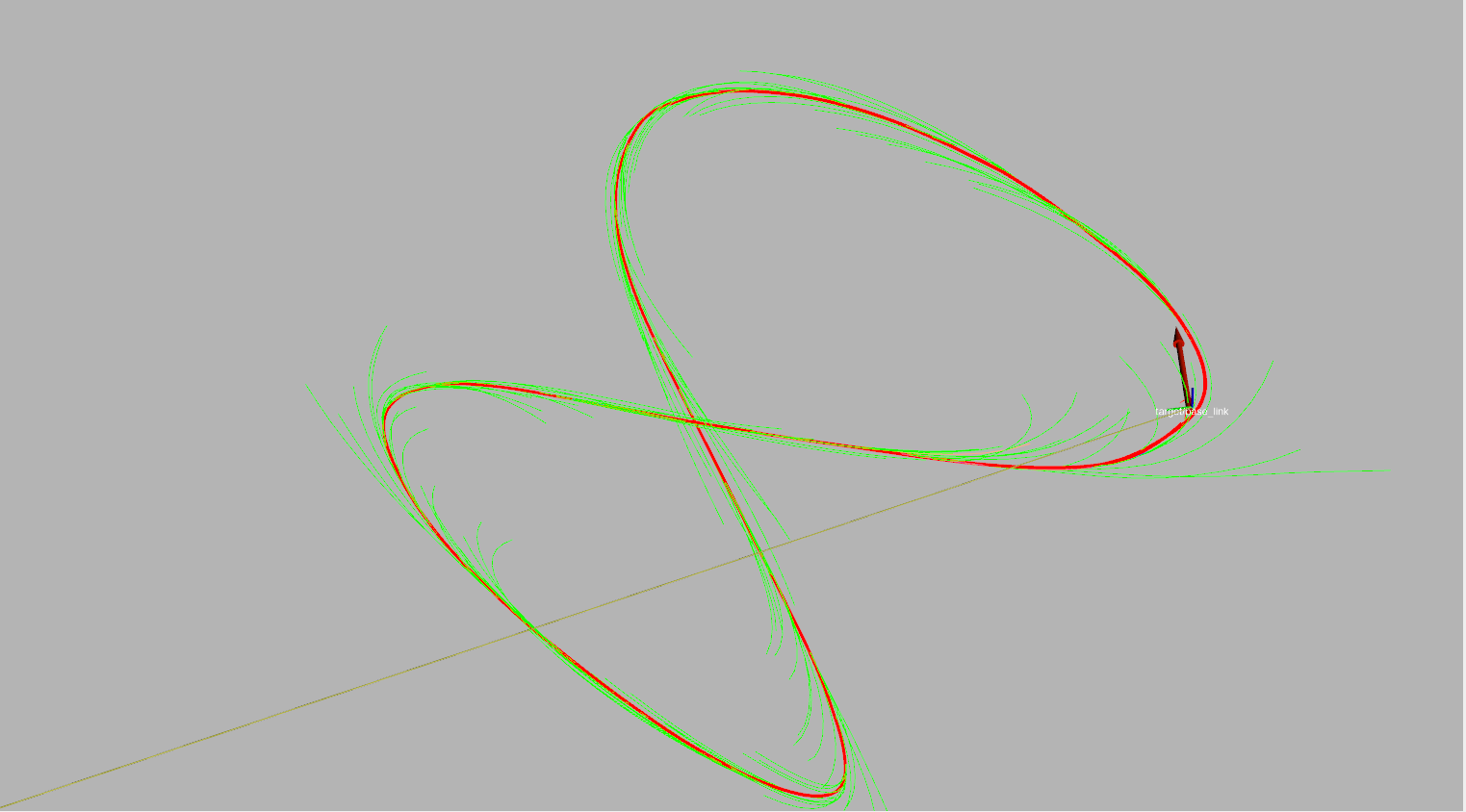}
  \caption{Trajectory prediction based on velocity model with whitening normalization.}
  \label{fig:sim_vel_max_norm}
\end{subfigure}%

\caption{Visualization of real-time position prediction in Gazebo simulation. The red path is the actual quadcopter trajectory. The green paths are the different predictions of position segments at different time instances. The quadcopter was flying in positions outside the position distribution of training dataset.}
\label{fig:sim_position_trajectories}
\end{figure*}

We conducted 4 simulation experiments using 4 models: position model with max normalization method, position model with whitening normalization method, velocity model with max normalization method, and velocity model with whitening normalization method. We report the average RMSE for each experiment in Table \ref{tab:sim_rmse}, and a visualization of the corresponding trajectory and the predictions are depicted in Figure \ref{fig:sim_position_trajectories}. 

we intentionally designed the trajectory to be outside the position ranges (position distribution) that are present in the training dataset. This is to evaluate the generalizability of the position and velocity models. As seen in Figure \ref{fig:sim_pos_max_norm} and \ref{fig:sim_pos_white_norm}, the predicted trajectories (in green) by the position models are not usable as they diverge from the actual trajectory (in red). Their reported average RMSE are very high, 46.241 and 19.73 for the position models using the max  and whitening normalization methods, respectively. On the other hand, position predictions based on the velocity models show much higher accuracy as in Figure \ref{fig:sim_vel_max_norm} and \ref{fig:sim_vel_white_norm}, where the predictions are close to the actual trajectory. The reported RMSE are 0.424 and 0.349 for the velocity models using the whitening and max normalization methods, respectively. This shows that the position prediction based on the velocity model can generalize better in unseen position distributions as they do not depend on the absolute position used in the training and their statistics.

\subsection{Real-world Experiments}

In this section, we demonstrate the power of the simulation-to-reality (sim2real) concept through real-world experiments. We used the same four models that were tested in the simulation experiments (see Section \ref{sec:sim_results}), which were trained exclusively on simulated trajectories within the ROS-Gazebo framework. These models show high quality trajectory predictions when applied in actual experiments.

The hardware setup primarily consists of the same quadcopter model used in the simulation, the X500 V2 shown in Figure \ref{fig:HolybroQuadcopter}, equipped with an onboard Pixhawk 6c flight controller running PX4 firmware. The flight controller is interfaced with an onboard computer, a Jetson Orin Xavier NX with 16GB RAM, which runs ROS 2 to communicate with the PX4 autopilot. The estimated states of the quadcopter, such as position and velocity, are received and fed into the trajectory prediction models, also running on the Jetson device.

The prediction models were running at a frequency of 20Hz on the Jetson without any hardware acceleration. It is expected that this frequency could significantly increase—potentially by 3 to 5 times—after converting the models from the current PyTorch format to TensorRT format, which allows for optimized execution.

Unlike the simulation experiments where the quadcopter followed a predefined lemniscate trajectory, in the real-world tests, the quadcopter was flown manually by a pilot performing arbitrary maneuvers at varying speeds. This allowed us to evaluate the prediction quality for maneuvers not explicitly represented in the training dataset. The prediction models were assessed using the same trajectory.

\begin{figure*}[ht]
\centering
\begin{subfigure}{.45\textwidth}
  \centering
  \includegraphics[width=.95\linewidth]{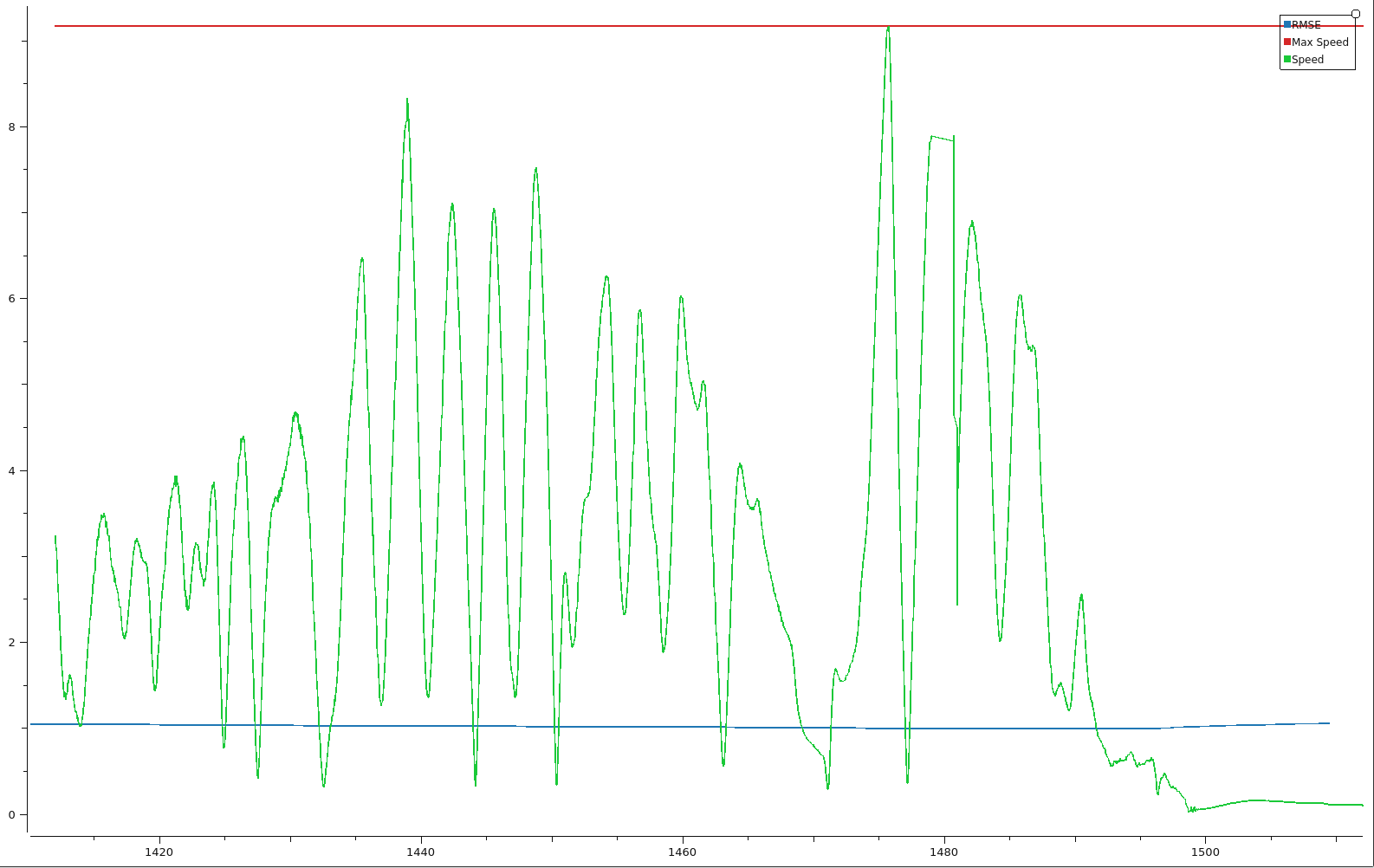}
  \caption{Results using position prediction model with whitening normalization.}
  \label{fig:real_pos_white_norm}
\end{subfigure}%
\begin{subfigure}{.45\textwidth}
  \centering
  \includegraphics[width=.95\linewidth]{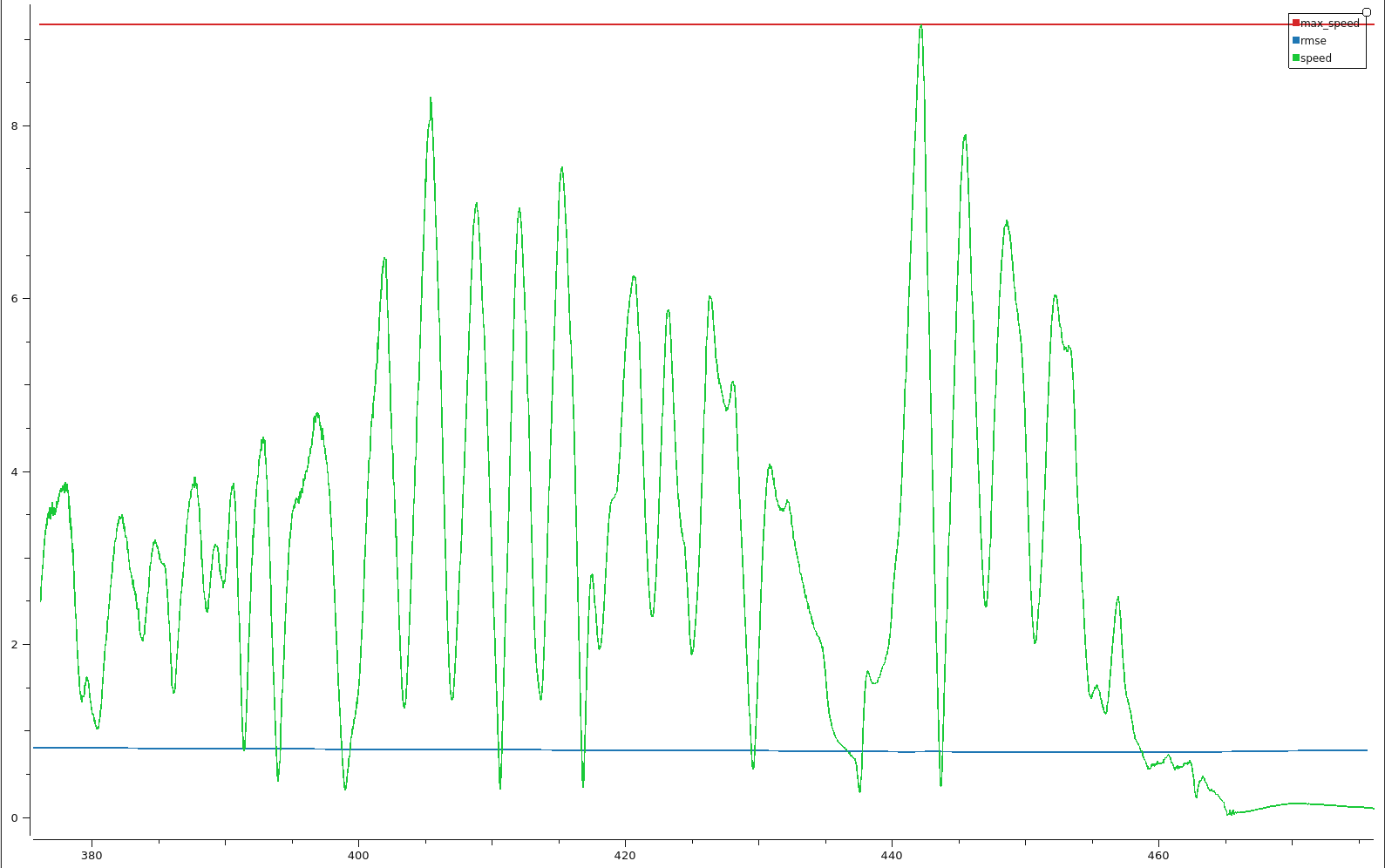}
  \caption{Results using position prediction model with maximum norm.}
  \label{fig:real_pos_max_norm}
\end{subfigure}%
\\ 

\begin{subfigure}{.45\textwidth}
  \centering
  \includegraphics[width=.95\linewidth]{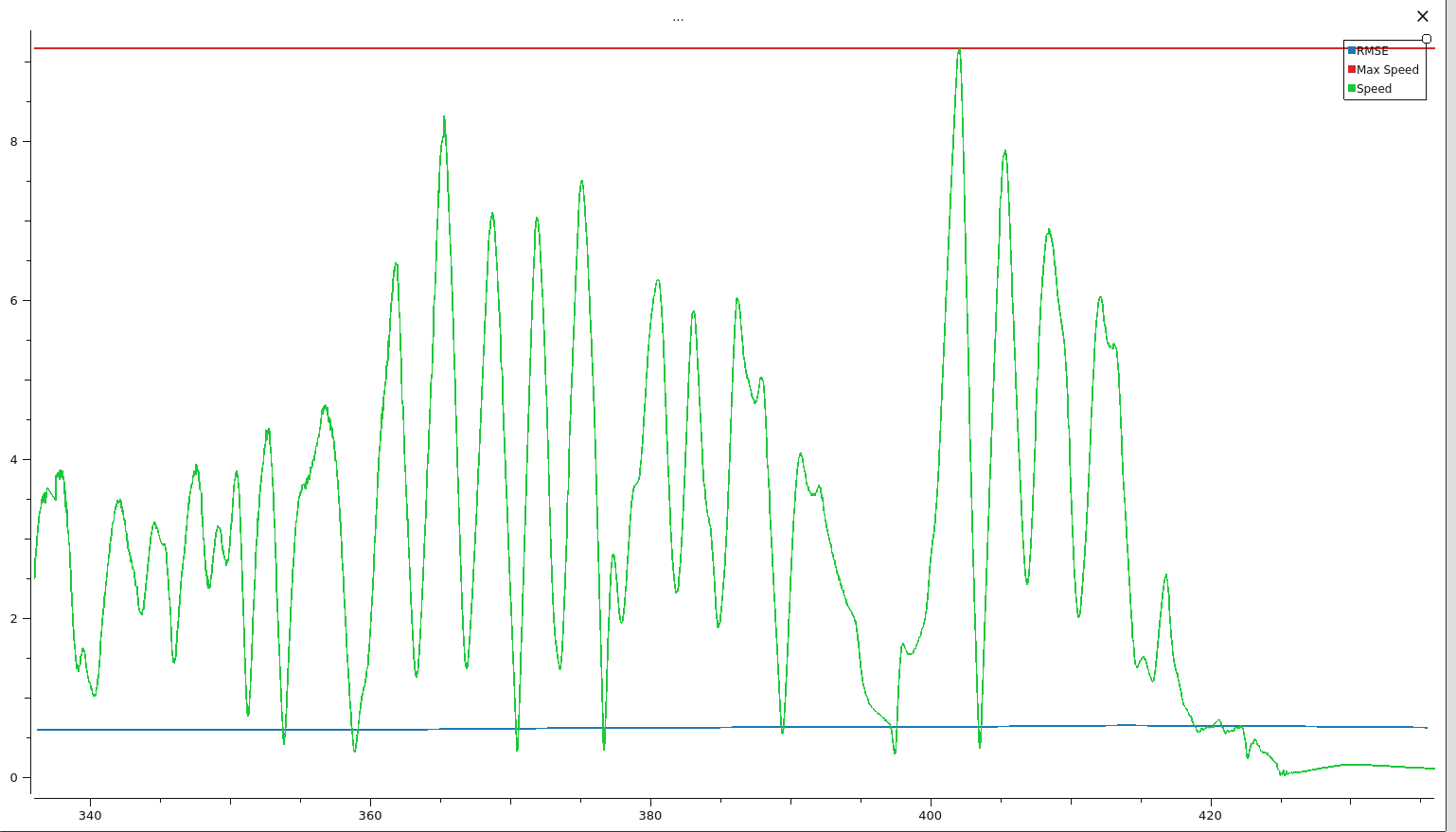}
  \caption{Results using velocity prediction model with whitening normalization.}
  \label{fig:real_vel_white_norm}
\end{subfigure}%
\begin{subfigure}{.45\textwidth}
  \centering
  \includegraphics[width=.95\linewidth]{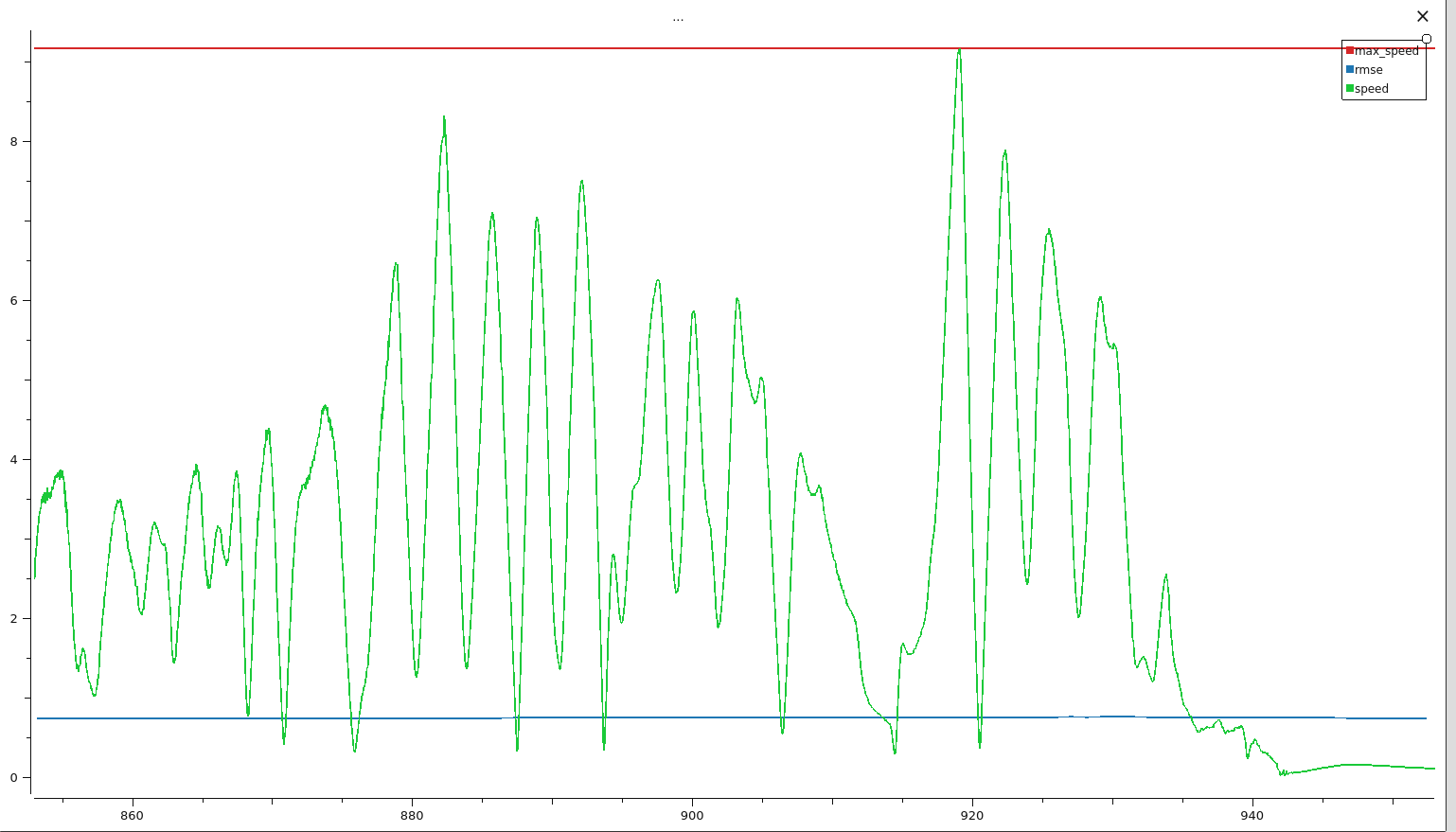}
  \caption{Results using velocity prediction model with maximum norm.}
  \label{fig:real_vel_max_norm}
\end{subfigure}%

\caption{Accuracy of 4 prediction models in real experiments. The horizontal axis represent time in seconds. The vertical axis represents RMSE (blue) in meters, speed (green) in m/s, maximum speed (red) in m/s.}
\label{fig:real_experiemtsn_performance}
\end{figure*}

In Fig. \ref{fig:real_experiemtsn_performance}, we show the real-time accuracy of the prediction models in terms of RMSE versus time, alongside the quadcopter’s speed, which ranged from 0 to 9.17 m/s. The average RMSE for all models is reported in Table \ref{tab:real_experiments_rmse}. Notably, the position predictions from the velocity-based models consistently outperformed those from the position-based models. The best position prediction accuracy, achieved with the velocity model using whitening normalization, recorded an RMSE of 0.622, while the position model with maximum norm had an RMSE of 0.776. Additionally, the models exhibited nearly constant RMSE values, regardless of speed variation during flight, indicating robustness to changes in quadcopter velocity.

\begin{table*}[ht]
\centering
\caption{Average RMSE of position and velocity-based trajectory prediction in real experiments}
\label{tab:real_experiments_rmse}
\resizebox{\textwidth}{!}{
\begin{tabular}{|c|c|}
\hline
Model & Average RMSE \\ \hline
Position prediction with whitening normalization in Figure \ref{fig:real_pos_white_norm} & 1.028 \\ \hline
Position prediction with maximum norm Figure \ref{fig:real_pos_max_norm}& 0.776 \\ \hline
Position prediction based on velocity model with maximum norm Figure \ref{fig:real_vel_max_norm}& 0.712 \\ \hline
Position prediction based on velocity model with whitening normalization Figure \ref{fig:real_vel_white_norm} & 0.622 \\ \hline
\end{tabular}
}
\end{table*}

\section{Conclusions}\label{sec:conclusions}

This paper presents a novel GRU-based solution for predicting future UAV positions by leveraging both position and velocity historical data. The GRU model’s efficiency in handling trajectory data makes it a robust alternative to more complex architectures like transformers and LSTMs. The proposed model was evaluated across 24 configurations using both synthetic and real-world UAV trajectories. Real-world experiments with a quadcopter performing arbitrary flight patterns confirmed the high accuracy of velocity-based predictions, outperforming position-based models and traditional RNNs. The GRU-based model consistently achieved low MSE values, ranging from $2 \times 10^{-8}$ to $2 \times 10^{-7}$, in both simulated and real-world conditions. These findings demonstrate the model’s applicability for real-time UAV prediction tasks and lay the foundation for extending this approach to other domains where accurate future location prediction is critical.

\section*{Acknowledgments}
This work is supported by \emph{the Robotics and Internet of Things Lab} of \emph{Prince Sultan University.}

\section*{Declaration of competing interest}
The authors declare that they have no known competing financial interests or personal relationships that could have appeared to influence the work reported in this paper.

\section*{Declaration of generative AI and AI-assisted technologies in the writing process}
During the preparation of this work the author(s) only used Grammarly and GPT-4 in order to improve language and readability. The authors have not used generative AI for any scientific analysis or interpretations. After using this tool/service, the author(s) reviewed and edited the content as needed and take(s) full responsibility for the content of the publication.

\section*{Data availability}
The data used in this study is available at: \url{https://huggingface.co/datasets/riotu-lab/Synthetic-UAV-Flight-Trajectories}.

\bibliographystyle{unsrtnat}
\bibliography{bib/references}

\end{document}